\documentclass{article}

\usepackage{arxiv}

\usepackage[utf8]{inputenc} 
\usepackage[T1]{fontenc}    
\usepackage{hyperref}       
\usepackage{url}            
\usepackage{booktabs}       
\usepackage{amsfonts}       
\usepackage{nicefrac}       
\usepackage{microtype}      
\usepackage{lipsum}
\usepackage{graphicx}
\graphicspath{ {./images/} }

\usepackage{textcomp}
\usepackage{xcolor}
\usepackage{graphicx}
\usepackage{color,soul}
\usepackage[shortlabels]{enumitem}
\usepackage{amsmath}
\usepackage{subcaption}

\title{Low-Dimensional State and Action Representation Learning with MDP Homomorphism Metrics}

\author{
 Nicolò Botteghi \\
  Robotics and Mechatronics\\
  University of Twente\\
  Enschede, The Netherlands \\
  \texttt{n.botteghi@utwente.nl} \\
  \And
 Mannes Poel \\
  Datamanagement and Biometrics\\
   University of Twente\\
  Enschede, The Netherlands \\
  \texttt{m.poel@utwente.nl} \\
  \And
  Beril Sirmacek \\
  Department of Smart Cities \\ 
  Saxion University of Applied Sciences \\
  \texttt{b.sirmacek@saxion.nl} \\
  \And
   Christoph Brune \\
  Applied Analysis\\
   University of Twente\\
  Enschede, The Netherlands \\
  \texttt{c.brune@utwente.nl} \\
}

\begin{document}
\maketitle
\begin{abstract}
Deep Reinforcement Learning has shown its ability in solving complicated problems directly from high-dimensional observations. However, in end-to-end settings, Reinforcement Learning algorithms are not sample-efficient and requires long training times and quantities of data. 
In this work, we proposed a framework for sample-efficient Reinforcement Learning that take advantage of state and action representations to transform a high-dimensional problem into a low-dimensional one. Moreover, we seek to find the optimal policy mapping latent states to latent actions. Because now the policy is learned on abstract representations, we enforce, using auxiliary loss functions, the lifting of such policy to the original problem domain. Results show that the novel framework can efficiently learn low-dimensional and interpretable state and action representations and the optimal latent policy.
\end{abstract}


\section{Introduction}
In the last decade, Deep Reinforcement Learning \cite{sutton_reinforcement_2018} algorithms have solved increasingly complicated problems in many different domains, spanning from video games \cite{mnih2013playing} to numerous robotics applications \cite{kober2013reinforcement}, in an end-to-end fashion. Despite the success of end-to-end Reinforcement Learning, these methods suffer from low sample efficiency and usually requires lengthy and expensive training procedures to learn optimal behaviours. This problem is even more emphasized when the true state of the environment is not observable, and the observation space $\mathcal{O}$ or the action space $\mathcal{A}$ are high-dimensional. In end-to-end settings, due to the weak supervision of the reward signal, Reinforcement Learning algorithms are not enforced to learn good state representations of the environment, making the mapping observations to actions challenging to learn and interpret.


State representation learning \cite{SRL_survey} methods aim at reducing the dimensionality of the observation stream by learning a mapping from the observation space $\mathcal{O}$ to a lower-dimensional state space $\mathcal{\Bar{S}}$ containing only the meaningful feature needed for solving a given task. By employing self-supervised auxiliary losses, it is possible to enforce optimal state representation and learn models of the underlying Markov Decision Process, or MDP. When policies are learned using the \textit{abstract} or \textit{latent} state-space variables, the training time is often reduced, the sample-efficiency, the robustness, and generalisation capabilities of the policies grow compared to end-to-end Reinforcement Learning \cite{gelada2019deepmdp}, \cite{francoislavet2018combined} and \cite{VanderPol2020}.

While the problem of state representation and observation compression has been extensively treated \cite{SRL_survey}, only a few works have extended the concept of dimensionality reduction to the action space $\mathcal{A}$. In this category, we find the works done in \cite{dulacarnold2015deep}, \cite{losey2019controlling} and \cite{ch2019learning} where low-dimensional action representations are used to improve training efficiency of the agents. In particular, the methods proposed in \cite{losey2019controlling} and \cite{ch2019learning} learn an action representation using self-supervised approaches. 

\begin{figure}[h!]
    \centering
    \includegraphics[width=0.75\textwidth,page=1]{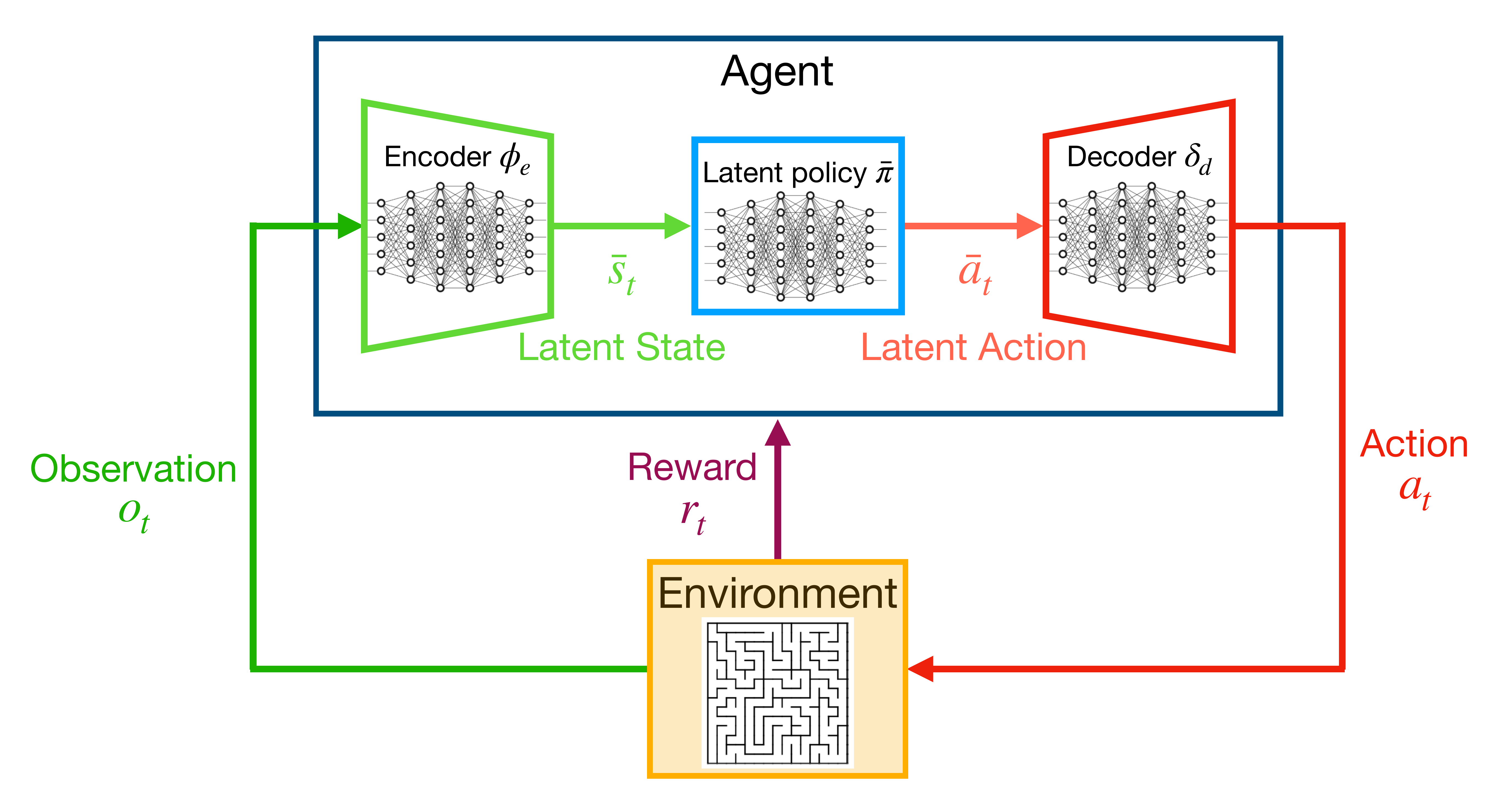}
    \caption{Proposed framework combining state and action representation with reinforcement learning.}
    \label{fig:state_action_representation_framework}
\end{figure}

In this paper, we study the problem of learning state and action representations, in the context of reinforcement learning. In particular, with reference to Figure \ref{fig:state_action_representation_framework}, we propose a unified framework  composed  of:
\begin{itemize}
    \item an encoder neural network $\phi_e$ mapping observations to low-dimensional latent states,    trained by leveraging on the knowledge of MDP homomorphism. In this way, we can have guarantees on the optimality of the policy learned using the latent state space $\bar{\mathcal{S}}$.
    \item the learning of a latent continuous policy $\bar{\pi}$ mapping latent states $\Bar{s} \in \mathcal{\Bar{S}}$ to latent actions $\Bar{a} \in \mathcal{\Bar{A}}$.
    \item the learning of a deterministic action decoder $\delta_d$, mapping then continuous latent action space $\mathcal{\Bar{A}}$ to the original action space $\mathcal{A}$. 
\end{itemize}
Because the optimal latent policy is learned using state and action representations, it is important to study if such a policy can be lifted to the original problem while preserving its optimality. For this purpose, we employ the notion of MDP homomorphism \cite{ravindran2001symmetries}, \cite{ravindran2004approximate}.

The rest of the paper is organized as follows: Section \ref{sec:Background} introduces the background information to this research, while Section \ref{sec:RelatedWork} presents the related work in the context on state and action representation for Reinforcement Learning. Section \ref{sec:Methodology} explains the methodology, and Section \ref{sec:ExperimentalDesign} the experimental design. Eventually, the results are presented in Section \ref{sec:Results}, followed by the discussion of the findings in Section \ref{sec:Discussion}, and conclusion in Section \ref{sec:Conclusion}.

\section{Background}\label{sec:Background}
\subsection{Reinforcement Learning}\label{subsec:RL}

A Markov Decision Process $\mathcal{M}$, or MDP, is a tuple $\langle \mathcal{S}, \mathcal{A}, \text{T}, \text{R} \rangle$, where $\mathcal{S}$ is the set of states of the environment, $\mathcal{A}$ is the set of actions that the agent can take, $ \text{T}: \mathcal{S} \times \mathcal{A} \times \mathcal{S} \longrightarrow [0,1]$  is the transition function mapping the current state and an action to the probability of transitioning to a next state state and $\text{R}:  \mathcal{S} \times \mathcal{A} \longrightarrow \mathbb{R}$ is the reward function assessing the quality of the agent's actions in all the states. The agent's goal is to find the best acting strategy, i.e. the optimal policy $\pi^*$, maximising the cumulative reward:
\begin{equation}
G_t = \sum_{t=0}^\infty \gamma^t\text{R}(s_t,a_t)
    \label{return}
\end{equation}
 where  $\gamma$ is the discount factor.
The expected return of a given state under a policy $\pi$ is  computed using the so-called value function $\text{V}^{\pi}: \mathcal{S} \longrightarrow \mathbb{R}$:
\begin{equation}
\text{V}^{\pi}(s) = \mathbb{E}_{\pi}[G_T|s_t=s]
    \label{value_function}
\end{equation}
where $\mathbb{E}_{\pi}$ is the expectation under the policy $\pi$, $G_t$ is the total return collected by the policy $\pi$, when the initial state is $s$.
Equivalently, we can estimated the expected return of a given state-action pair  under a policy $\pi$ using the action-value function $\text{Q}^{\pi}: \mathcal{S} \times \mathcal{A} \longrightarrow \mathbb{R}$:
\begin{equation}
\text{Q}^{\pi}(s,a) = \mathbb{E}_{\pi}[G_T|s_t=s,a_t=a]
    \label{action-value_function}
\end{equation}
where $\mathbb{E}_{\pi}$ is the expectation under the policy $\pi$, $G_t$ is the total return collected by the policy $\pi$ when the initial state is $s$ and the action taken is $a$.

In many scenarios, the agent has no knowledge of the environment dynamics, i.e. the transition and reward functions are unknown. Therefore, dynamic programming algorithms, such as Value Iteration \cite{sutton_reinforcement_2018}, that exploit the MDP models for planning cannot be directly applied. However, in the context, Reinforcement Learning \cite{sutton_reinforcement_2018} can be employed. Reinforcement Learning agents aim at learning the optimal policies by only using the sample tuples $(s,a,r,s')$ collected through the interaction with the environment.

\subsection{MDP homomorphism}\label{subsec:MDP-homomorphism}

When learning representations of the original MDP, we would like to preserve its structure to lift the optimal policies, learned given the representations, to the original MDP by preserving its optimality. This can be done by the notion of MDP homomorphism \cite{ravindran2001symmetries}, \cite{ravindran2004approximate}.

\textbf{\textit{Definition:} } (Adapted from \cite{ravindran2004approximate}) A stochastic MDP homomorphism $h$ from an MDP $\mathcal{M} = \langle \mathcal{S}, \mathcal{A}, \text{T}, \text{R}  \rangle$ to an MDP $\mathcal{\Bar{M} = \langle \Bar{S}, \Bar{A}}, \Bar{\text{T}}, \Bar{\text{R}}  \rangle$ is a tuple $\langle f, g_s \rangle$, with:
\begin{itemize}
    \item $f: \mathcal{S} \longrightarrow \mathcal{\Bar{S}}$
    \item $g_s: \mathcal{A} \longrightarrow \mathcal{\Bar{A}}$
\end{itemize}
such that the following identities hold:
\begin{equation}
    \forall_{s, s' \in \mathcal{S}, a \in \mathcal{A}} \ \ \ \Bar{\text{T}}(f(s')|f(s), g_s(a)) = \sum_{s'' \in {[s']_f}} \text{T}(s''|s,a)
    \label{transition_mdpHomo_st}
\end{equation}
\begin{equation}
    \forall_{s, a \in \mathcal{A}} \ \ \ \Bar{\text{R}}(f(s), g_s(a)) = \text{R}(s, a)
    \label{reward_mdpHomo_st}
\end{equation}
where $[s']_f$  is the equivalence class of $s'$ under Z.

\textbf{\textit{Definition:} } (Adapted from \cite{VanderPol2020}): A deterministic MDP homomorphism $h$ from an MDP $\mathcal{M} = \langle \mathcal{S}, \mathcal{A}, T, R  \rangle$ to an MDP $\mathcal{\Bar{M} = \langle \Bar{S}, \Bar{A}}, \Bar{\text{T}}, \Bar{\text{R}}  \rangle$ is a tuple $\langle f, g_s \rangle$, with:
\begin{itemize}
    \item $f: \mathcal{S} \longrightarrow \mathcal{\Bar{S}}$
    \item $g_s: \mathcal{A} \longrightarrow \mathcal{\Bar{A}}$
\end{itemize}
such that the following identities hold:
\begin{equation}
    \forall_{s, s' \in \mathcal{S}, a \in \mathcal{A}} \ \ \ \text{T}(s,a) = s' \Longrightarrow \Bar{\text{T}}(f(s), g_s(a)) = f(s')
    \label{transition_mdpHomo}
\end{equation}
\begin{equation}
    \forall_{s \in \mathcal{S}, a \in \mathcal{A}} \ \ \ \Bar{\text{R}}(f(s), g_s(a)) = \text{R}(s, a)
    \label{reward_mdpHomo}
\end{equation}

If Equation (\ref{transition_mdpHomo_st}), (\ref{reward_mdpHomo_st}) or (\ref{transition_mdpHomo}), (\ref{reward_mdpHomo}) are satisfied, the optimal policy $\bar{\pi}$ of the homomorphic image $\bar{\mathcal{M}}$ can be lifted to the original MDP $\mathcal{M}$. Therefore, for deterministic policies $\pi(s)$ and $\bar{\pi}(f(s))$ and a deterministic mapping $f$, we can write:
\begin{equation}
    \pi(s) = \bar{\pi}(f(s))
\end{equation}

In this work, we focus on the case of deterministic MDPs (all the details in Section \ref{sec:Methodology}) with deterministic transition function $\text{T}:\mathcal{S} \times \mathcal{A} \longrightarrow \mathcal{S}$, state space $\mathcal{S}$ not observable, but Markovian observation space $\mathcal{O}$. With reference to Figure \ref{fig:computational_scheme_framework}, we define $f$ as the observation encoder $\phi_e:  \mathcal{O}  \longrightarrow \bar{\mathcal{S}}$ mapping observations to latent states, and $g_s$ as the function $\psi_e: \bar{\mathcal{S}} \times \mathcal{A} \longrightarrow \bar{\mathcal{A}}$ mapping latent states and actions to latent actions.

\subsection{Twin Delayed Deep Deterministic Policy Gradient}\label{subsec:TD3}

The proposed approach transforms the observation and action spaces into continuous latent state and action spaces. Therefore the optimal latent policy $\bar{\pi}^*$ we aim to find is necessarily a continuous policy mapping latent states to latent actions. Differently from \cite{VanderPol2020}, we do not employ any discretisation of the latent state space that would limit the applicability of our framework to problems with continuous state and action spaces.  

We employ the Twin Delayed Deep Deterministic Policy Gradient \cite{fujimoto2018addressing}, or TD3, algorithm\footnote{In principle, any Reinforcement Learning algorithm which is suitable for continuous state and action spaces can be used.}. Inspired by Double Deep-Q Network \cite{van2016deep}, or DDQN, TD3 addresses the problem of  overestimation of the action-value function of Deep Deterministic Policy Gradient \cite{lillicrap2015continuous}, or DDPG. To prevent the overestimation of the action-value function, TD3 utilizes two critic neural networks estimating the action-value function $\hat{\text{Q}}_1(s,a;\pmb{\theta}_{\hat{\text{Q}}_1})$ and $\hat{\text{Q}}_2(s,a;\pmb{\theta}_{\hat{\text{Q}}_2})$, parametrised by ${\pmb{\theta}_{\hat{\text{Q}}_1}}$ and  ${\pmb{\theta}_{\hat{\text{Q}}_2}}$ respectively, and an actor neural network $\pi(s;\pmb{\theta}_{\pi})$ approximating a continuous policy, parametrised by $\pmb{\theta}_{\pi}$. The policy $\pi_{\pmb{\theta}_{\pi}}: \mathcal{S} \longrightarrow \mathcal{A}$ is a deterministic policy mapping states to actions, but to guarantee sufficient exploration during the training phase, noise $\epsilon \sim \mathcal{N}(0, \sigma)$, from a Gaussian distribution $\mathcal{N}(0, \sigma)$ with zero mean and standard deviation $\sigma$, is added to the action $a \sim \pi(s;\pmb{\theta}_{\pi}) + \epsilon$. As in any other deep reinforcement learning algorithm, the experience tuples $(s,a,r,s')$, collected  through the interaction with the environment, are stored in the memory buffer and used to update the neural network. Similarly to DDPG, TD3 uses target networks for critics $\hat{\text{Q}}_1^-$, $\hat{\text{Q}}_2^-$  and actor $\pi^-$ with parameters ${\pmb{\theta}_{\hat{\text{Q}}_1^-}}$, ${\pmb{\theta}_{\hat{\text{Q}}^-_2}}$, and ${\pmb{\theta}_{\pi^-}}$ respectively. 


While in DDPG, the Temporal Difference, or TD, error $y$ is computed by the target critic network (see Equation (\ref{TDerrorDDPG})), TD3 makes use of both critic networks to reduce the overestimation generated by the use of a single value function estimator, as it can be seen in Equation (\ref{TDerrorTD3}).
\begin{equation}
    y =  r + \gamma \hat{\text{Q}}^-(s',a;\pmb{\theta}_{\hat{\text{Q}}^-})
    \label{TDerrorDDPG}
\end{equation}
\begin{equation}
    y =  r + \gamma \min_{i=1,2}\hat{\text{Q}}^-_i(s',\Tilde{a};\pmb{\theta}_{\hat{\text{Q}}_i^-})
    \label{TDerrorTD3}
\end{equation}
where $\Tilde{a} \sim \pi^-(s;\pmb{\theta}_{\pi^-}) +  \epsilon$, and $\epsilon \sim \text{clip}(\mathcal{N}(0,\Tilde{\sigma}),-c,c)$, $c$ a constant hyperparameter of the algorithm.
The TD-error is then used to generate a fixed target for the training of the critic networks and their parameters' update, as shown in Equation (\ref{critics_update}).
\begin{equation}
   {\pmb{\theta}_{\hat{\text{Q}}_i}} \longleftarrow \min_{\pmb{\theta}_{\hat{\text{Q}}_i}} \frac{1}{N} \sum (y-\hat{\text{Q}}_i(s,a))^2
    \label{critics_update}
\end{equation}
As in DDPG, the actor is updated using the deterministic policy gradient theorem and the gradient of the critic $\nabla_a \hat{\text{Q}}_1(s,a)$ as shown in Equation (\ref{actor_update}). However, in TD3, the actor is updated with a lower frequency than the critics.
\begin{equation}
   \nabla_{\pmb{\theta}_{\pi}} J(\pmb{\theta}_{\pi}) = \frac{1}{N} \sum  \nabla_a \hat{\text{Q}}_1(s,a)|_{a=\pi(s)} \nabla_{\pmb{\theta}_{\pi}} \pi(s) 
    \label{actor_update}
\end{equation}
Eventually, the  target networks are updated, as in:
\begin{equation}
    \begin{split}
       {\pmb{\theta}_{\hat{\text{Q}}^-_i}}  &\longleftarrow \tau {\pmb{\theta}_{\hat{\text{Q}}_i}} + (1-\tau){\pmb{\theta}_{\hat{\text{Q}}^-_i}} \\
      {\pmb{\theta}_{\pi^-}}  &\longleftarrow \tau {\pmb{\theta}_{\pi}} + (1-\tau){\pmb{\theta}_{\pi^-}} \\
    \end{split}
\end{equation}
where $\tau$ is the hyperparameter controlling the  speed of the updates.



\section{Related Work}\label{sec:RelatedWork}
\subsection{Learning State Abstractions} 

The notion of MDP homomorphism was first introduced in \cite{ravindran2001symmetries}, \cite{ravindran2004approximate}, \cite{taylor2008bounding} for exploiting symmetries and similiaties in MDPs and minimise their models. 
In more recent year, the MDP homomorphism metrics were used for learning low-dimensional state representation in the context of Reinforcement Learning \cite{francoislavet2018combined}, \cite{gelada2019deepmdp}, \cite{VanderPol2020}. The MDP  homomorphism metrics are used as auxiliary loss functions for training neural networks. When such losses approach zero, it is possible to prove that we have found a homomorphic image of the original MDP. When learning the observation to latent states mapping, a contrastive loss \cite{VanderPol2020}, \cite{francoislavet2018combined} is necessary for preventing the collapse of the mapping. This problem is frequent when the reward function is space \cite{gelada2019deepmdp}. 


Many other state representation learning approaches for Reinforcement Learning have been proposed in literature \cite{SRL_survey} and most of them employ Auto-Encoder, or AE, reconstruction losses to learn the mapping to the latent state. However, in the context of Reinforcement Learning, where the main goal is only to use the latent state information, the reconstructed observations are usually discarded by making the decoder a non-required and additional complexity. Moreover, AE-based methods tend to struggle to encode and reconstruct non-salient features, and they are easily "distracted" textures or background features. This means that the encoder cannot select between salient and relevant-to-the-task features. 

To overcome this problem, several approaches associate to the AE loss, a latent transition loss, or a reward loss, an inverse model loss or a combination of those \cite{finn2016deep}, \cite{mattner2012learn}, \cite{van2016stable}, \cite{de2018integrating}, \cite{wahlstrom2015pixels}.  In \cite{Francois-Lavet2019}, a framework combining model-free and model-based RL based on the learning of a state representation using multi-objective loss function is proposed. The collapsing of the state representation due to sparse rewards is tackled by using two contrastive losses. 

Eventually, we can find approaches for state representation learning that utilize prior knowledge to shape the latent state space through auxiliary loss functions \cite{jonschkowski2015learning}, \cite{jonschkowski2017pves}, \cite{botteghi2020low}. These methods have proven to be sample efficient and suitable for all the situations in which a low-date regime is required and especially useful in all the cases in which we have prior knowledge of the true environment space, e.g. in robotics where physical laws govern the true state space.

\subsection{Learning Action Abstractions}
Action abstraction in MDPs has been first introduced in \cite{sutton1999between} where a hierarchical decomposition of the policies is proposed to quickly learn skills and complicated tasks by simplifying the policies search space. Here, the low-level policies, i.e. the skill, are executed for a certain amount of steps. Only after their termination, the high-level policies are allowed to act and choose another skill.  Several methods have build upon the idea of temporal action abstraction namely the Option framework \cite{stolle2002learning}, the Max-Q \cite{dietterich1998maxq} and the Feudal networks \cite{vezhnevets2017feudal}.

In \cite{dulacarnold2015deep}, the authors utilise prior information over the action space to embed it in a low dimensional continuous space and allow the generalisation of RL algorithms when the original action space is highly discretised.
In \cite{ch2019learning}, the author proposed a method for exploiting the action structure by learning a decoder mapping from chosen low dimensional continuous action space, where the policy is learned, to the original action space. This work is the most related to ours. However, we look at the whole problem of state and action representation and their relation. 

\subsection{Learning State and Action Abstractions}

 Our work is related to \cite{pritz2020joint}, where state and action embeddings are learned in self-supervised settings for improving the performance of the Reinforcement Learning agent planning and acting in the learned embedding spaces. Differently, we do not assume full state observability, but we aim at learning a low-dimensional state representation from high-dimensional observations.
 

\section{Methodology}\label{sec:Methodology}

In this work, we study the interplay between state abstraction and action abstraction. While on one side, state representation learning allows reducing the dimensionality of the input space to exploit similarities and symmetries of the underlying (not-observable) true state space and speed up the learning of the policy and value function, we argue that an action representation should do the same. In particular, we aim at exploiting the underlying action space structure by representing it into a low-dimensional continuous space. 

\begin{figure}[h!]
    \centering
    \includegraphics[page=2, width=0.35\textwidth]{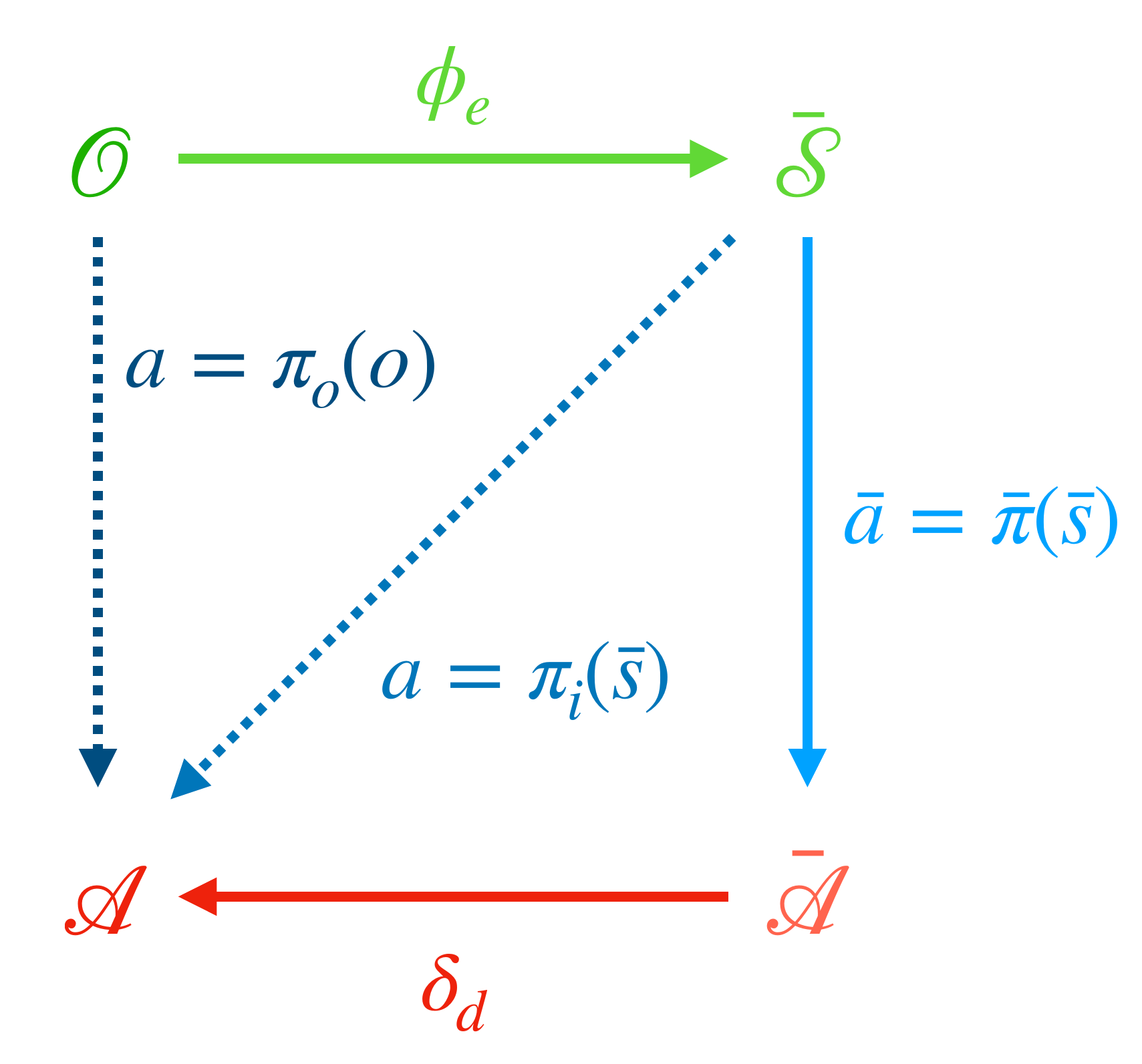}
    \caption{Relation between observation, action, latent state, and latent action spaces.}
    \label{fig:comparison_diagram}
\end{figure}

With reference to Figure \ref{fig:comparison_diagram}, we employ a state encoder neural network $\phi_e: \mathcal{O} \longrightarrow \mathcal{\Bar{S}}$, parametrised by $\pmb{\theta}_{\phi_e}$, mapping the observation space\footnote{We assume that the true state of the environment $\mathcal{S}$ is not directly observable by the agent. However, the  agent can perceive the world by means of high-dimensional observations. Similarly to other the work in \cite{VanderPol2020}, we restrict to the case of Markovian observations, i.e. a single observation contains enough  information for retrieving a good state representation.} $\mathcal{O}$ to a lower-dimensional latent state space $\mathcal{\Bar{S}}$, we learn a  continuous latent policy $\bar{\pi}: \bar{\mathcal{S}}  \longrightarrow \bar{\mathcal{A}}$, mapping latent states to latent actions, parametrised by a neural network with parameters $\pmb{\theta}_{\bar{\pi}}$, and eventually we map the latent actions back to the original action space $\mathcal{A}$ by means of a decoder $\delta_d: \mathcal{\Bar{A} \longrightarrow \mathcal{A}}$, parametrised by $\pmb{\theta}_{\delta_d}$. Morever, we indicate with $\pi_i: \bar{\mathcal{S}} \longrightarrow \mathcal{A}$ the policy mapping latent states to actions\footnote{We refer to the policy $\pi_i$ as the intermediate policy.},  and with $\pi_o: \mathcal{S} \longrightarrow \mathcal{A}$ the policy, mapping states to actions, of the original MDP.

The policy $\bar{\pi}$ is a latent policy, learned based on the latent state space $\bar{\mathcal{S}}$ and latent action space $\bar{\mathcal{A}}$, therefore, to guarantee its optimality and its lifting to the original state space $\mathcal{S}$ and action space $\mathcal{A}$, we use the notion of MDP homomorphism. Herein, we formally study under which conditions an optimal latent policy $\bar{\pi}^*$ is equivalent to the optimal intermediate policy $\pi_i^*$ and to the optimal original policy $\pi_o^*$.

The proposed framework is self-supervised and does not need labelled data. We only make full use of the experience tuple $(o,a,r,o')$ collected during the agent's interaction with the environment.


\subsection{State and Action Representation Learning}\label{subsec:LossFunctions}

\begin{figure*}[h!]
\centering
    \begin{subfigure}{0.49\textwidth}
    \centering
\includegraphics[page=1, width=0.6\linewidth]{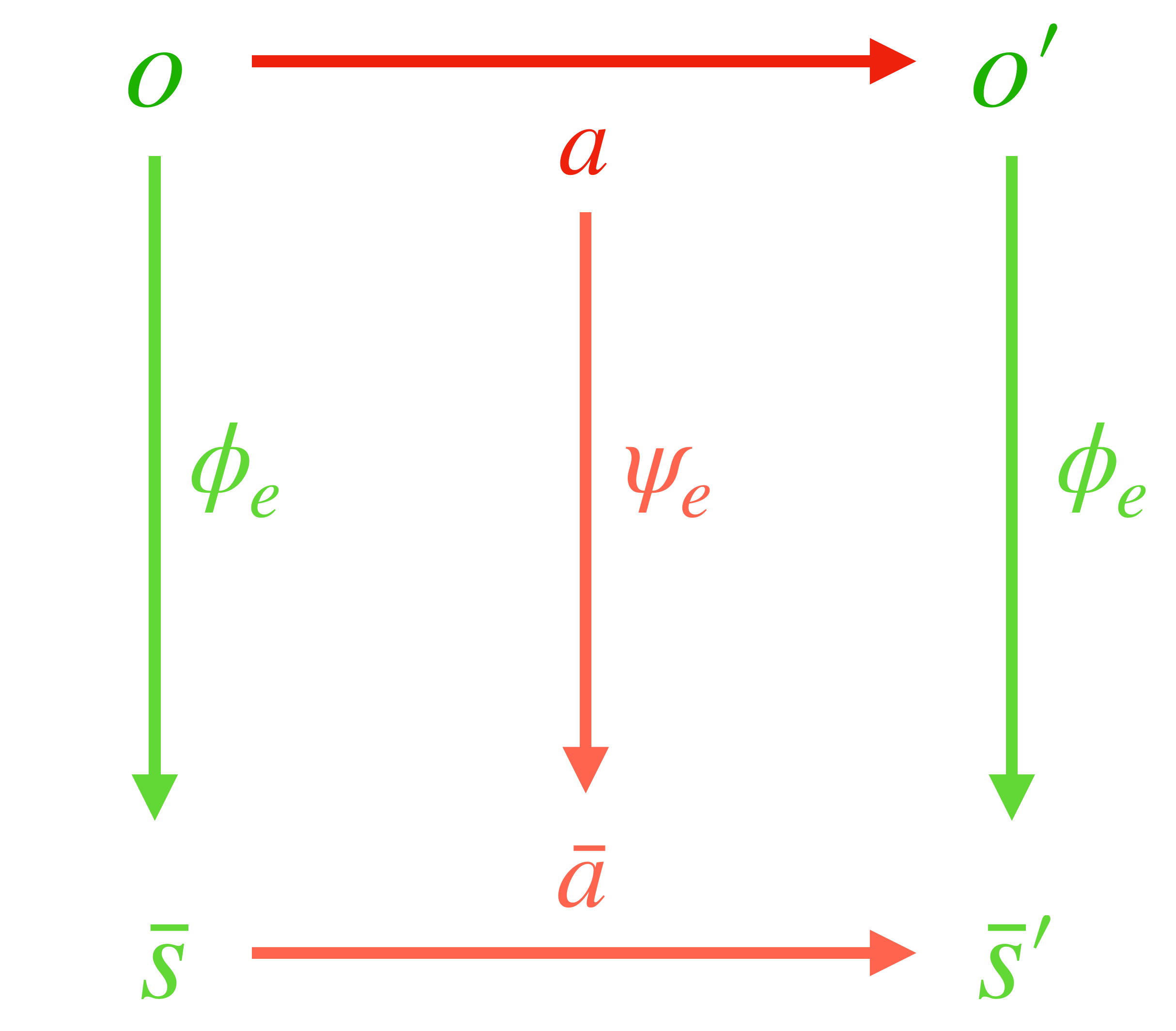}
\captionsetup{justification=centering}
    \caption{}
\label{fig:samples}
    \end{subfigure} 
    \begin{subfigure}{0.49\textwidth}
    \centering
\includegraphics[page=3, width=1.0\linewidth]{pics/networks3.pdf}
\captionsetup{justification=centering}
    \caption{}
\label{fig:models}
    \end{subfigure} 
\caption{Computational schemes of the proposed framework.}
    \label{fig:computational_scheme_framework}
\end{figure*}

The computational schemes of the proposed framework are presented in Figure \ref{fig:computational_scheme_framework}. Our approach combines two elements:
\begin{itemize}
    \item learning of a low-dimensional state representation $\bar{s}$ using the MDP homomorphism metrics
    \item learning of a low-dimensional action representation $\bar{a}$ to represent the action space $\mathcal{A}$  
\end{itemize}
such that, as shown in Figure \ref{fig:samples}, encoding through $\phi_e$ the next observation $o'$, obtained by applying action $a$ given the observation $o$, is equivalent to the encoding observation $o$ through $\phi_e$ and applying the latent action $\bar{a}$.

\subsubsection{Learning Low-dimensional State Representations}\label{subsubsec:StateEncoder}

We aim at learning an MDP homomorphism $\mathcal{\bar{M}}=\langle\mathcal{\bar{S}},\mathcal{\bar{A}},\bar{\text{T}}, \bar{\text{R}}\rangle$ of the original MDP $\mathcal{M}=\langle\mathcal{O},\mathcal{A},\text{T}, \text{R}\rangle$. Similarly to \cite{gelada2019deepmdp} and \cite{VanderPol2020}, the observation encoder $\phi_e$ is trained by means of the MDP homomorphism metrics, in Equation (\ref{transition_mdpHomo}) and (\ref{reward_mdpHomo}), and without the need of an observation decoder.

With reference to Figure \ref{fig:models}, we define a latent transition model $\Bar{\text{T}}: \mathcal{\Bar{S}} \times \mathcal{\Bar{A}} \longrightarrow \mathcal{\Bar{S}}$, predicting the next latent state given a latent state-action pair, and a latent reward model  $\Bar{\text{R}}: \mathcal{\Bar{S}} \times \mathcal{\Bar{A}} \longrightarrow \mathbb{R}$, predicting the reward of a latent state-action pair. Both mapping are learned with neural networks with parameters' vector $\pmb{\theta}_{\bar{\text{T}}}$ and $\pmb{\theta}_{\bar{\text{R}}}$ respectively.

Firstly, the transition loss, in Equation (\ref{transition_loss}), is used to enforce that transitions $ \text{T}(o,a)$ in original MDP $\mathcal{M}$ correspond to transitions $\bar{\text{T}}(\bar{s},\bar{a})$ in the latent  MDP $\mathcal{\bar{M}}$. Similarly to \cite{VanderPol2020}, \cite{kipf2019contrastive}, we model the transitions in the latent spaces as
$\bar{\text{T}}(\bar{s},\bar{a})=\Delta\bar{\text{T}}(\bar{s}, \bar{a}) + \bar{s}$.
\begin{equation}
\begin{split}
    \mathcal{L}_{\Bar{\text{T}}}(\pmb{\theta}_{\phi_e},\pmb{\theta}_{\bar{\text{T}}},\pmb{\theta}_{\psi_e}) &= \mathbb{E}[|| \bar{s}' - \hat{s}'||_2] \\
    &= \mathbb{E}[|| \bar{s}' - \bar{\text{T}}(\bar{s},\bar{a};\pmb{\theta}_{\bar{\text{T}}})||_2] \\
    &= \mathbb{E}[|| \bar{s}' - (\Delta\bar{\text{T}}(\bar{s},\bar{a};\pmb{\theta}_{\bar{\text{T}}})+\bar{s})||_2] \\
    &= \mathbb{E}[ \mid \mid \phi_e(o'; \pmb{\theta}_{\phi_e}) - (\Delta \bar{\text{T}}(\phi_e(o; \pmb{\theta}_{\phi_e}), \bar{a};\pmb{\theta}_{\bar{\text{T}}})+\phi_e(o;\pmb{\theta}_{\phi_e})) \mid \mid_2] \\
    &= \mathbb{E}[ \mid \mid \phi_e(o'; \pmb{\theta}_{\phi_e}) - (\Delta \bar{\text{T}}(\phi_e(o; \pmb{\theta}_{\phi_e}), \psi_e(\bar{s},a;\pmb{\theta}_{\psi_e});\pmb{\theta}_{\bar{\text{T}}})+\phi_e(o;\pmb{\theta}_{\phi_e})) \mid \mid_2] \\
\end{split}
\label{transition_loss}
\end{equation}
where the target next latent state $\bar{s}'=\phi_e(o'; \pmb{\theta}_{\phi_e})$ is generated by encoding the next observation $o'$, while the next latent state prediction $\hat{s}'=\bar{\text{T}}(\bar{s},\bar{a};\pmb{\theta}_{\bar{\text{T}}})$ is generated from the encoding $\bar{s}=\phi_e(o; \pmb{\theta}_{\phi_e})$ of the observation $o$, the action $a$, and the latent transition model $\bar{\text{T}}(\bar{s},\bar{a};\pmb{\theta}_{\bar{\text{T}}})$.

Secondly the reward loss, in Equation (\ref{reward_loss}), is used to enforce the same reward function in the original MDP $\mathcal{M}$ and the latent MDP $\mathcal{\bar{M}}$.
\begin{equation}
\begin{split}
\mathcal{L}_{\Bar{\text{R}}}(\pmb{\theta}_{\phi_e},\pmb{\theta}_{\bar{\text{R}}},\pmb{\theta}_{\psi_e}) &= \mathbb{E}[|| r - \hat{r}||_2]\\
&= \mathbb{E}[|| r - \Bar{\text{R}}(\bar{s},\bar{a};\pmb{\theta}_{\bar{\text{R}}})||_2]\\
&=\mathbb{E}[ \mid \mid r - \Bar{\text{R}}(\phi_e(o;\pmb{\theta}_{\phi_e}),\bar{a};\pmb{\theta}_{\bar{\text{R}}})\mid \mid_2] \\
&=\mathbb{E}[ \mid \mid r - \Bar{\text{R}}(\phi_e(o;\pmb{\theta}_{\phi_e}),\psi_e(\bar{s},a;\pmb{\theta}_{\psi_e});\pmb{\theta}_{\bar{\text{R}}})\mid \mid_2] \\
\end{split}
\label{reward_loss}
\end{equation}
where $r$ is the reward obtained by interacting with the environment and  $\hat{r}= \Bar{\text{R}}(\bar{s},\bar{a};\pmb{\theta}_{\bar{\text{R}}})$ is the predicted reward using the learned reward model $\bar{\text{R}}(\bar{s},\bar{a};\pmb{\theta}_{\bar{\text{R}}})$, the current observation $o$, and the action $a$.

Additionally, we used the hinge loss in Equation (\ref{contrastive_loss}) to prevent the trivial embedding in which all the latent states are mapped to the zero vector\footnote{This is often the case when the rewards are sparse \cite{VanderPol2020}, \cite{kipf2019contrastive}.} as this would not be an MDP homomorphism. 
\begin{equation}
\begin{split}
\mathcal{L}_c(\pmb{\theta}_{\phi_e},\pmb{\theta}_{\bar{\text{T}}},\pmb{\theta}_{\psi_e}) &= \mathbb{E}[\max(0, \epsilon - \mid \mid  \bar{s}_n' - \hat{s}' \mid \mid_2])\\
&= \mathbb{E}[\max(0, \epsilon - \mid \mid  \bar{s}_n' - \bar{\text{T}}(\bar{s},\bar{a};\pmb{\theta}_{\bar{\text{T}}}) \mid \mid_2])\\
&= \mathbb{E}[\max(0, \epsilon - \mid \mid  \phi_e(o_n';\pmb{\theta}_{\phi_e}) - (\Delta \bar{\text{T}}(\phi_e(o;\pmb{\theta}_{\phi_e}), \bar{a};\pmb{\theta}_{\bar{\text{T}}}) + \phi_e(o;\pmb{\theta}_{\phi_e})) \mid \mid_2 )] \\
&= \mathbb{E}[\max(0, \epsilon - \mid \mid  \phi_e(o_n';\pmb{\theta}_{\phi_e}) - (\Delta \bar{\text{T}}(\phi_e(o; \pmb{\theta}_{\phi_e}), \psi_e(\bar{s},a;\pmb{\theta}_{\psi_e});\pmb{\theta}_{\bar{\text{T}}})+\phi_e(o;\pmb{\theta}_{\phi_e})) \mid \mid_2 )] \\
\end{split}
\label{contrastive_loss}
\end{equation}
where $\epsilon$ is the hinge parameter governing the effect of the negative distance, and $o_n'$ is a randomly sampled observation, not a successor of the observation $o$.

The total loss for enforcing the MDP homomorphism is shown in Equation (\ref{total_loss_MDP}).
\begin{equation}
\mathcal{L}_{\text{MDP}} = \omega_{\Bar{\text{T}}}\mathcal{L}_{\Bar{\text{T}}}(\pmb{\theta}_{\phi_e},\pmb{\theta}_{\bar{\text{T}}},\pmb{\theta}_{\psi_e})  + \omega_{\Bar{\text{R}}}\mathcal{L}_{\Bar{\text{R}}}(\pmb{\theta}_{\phi_e},\pmb{\theta}_{\bar{\text{R}}},\pmb{\theta}_{\psi_e}) +\omega_c\mathcal{L}_c(\pmb{\theta}_{\phi_e},\pmb{\theta}_{\bar{\text{T}}},\pmb{\theta}_{\psi_e})
\label{total_loss_MDP}
\end{equation}
where $\omega_{\Bar{\text{T}}}, \omega_{\Bar{\text{R}}},$ and $\omega_c$ are three constants weighting the contribution of the each individual loss function.

\subsubsection{Learning Low-dimensional Action Representations}

Our second objective is to exploit similarities and structure of the action space $\mathcal{A}$. To do that we employ an action encoder $\psi_e:\bar{\mathcal{S}} \times \mathcal{A} \longrightarrow \bar{\mathcal{A}}$, mapping latent states and actions to state-dependent latent actions\footnote{The action encoder can be solely chosen a function of the actions $\psi_e:\mathcal{A}\longrightarrow\mathcal{\bar{A}}$.}, and an action decoder $\delta_d: \mathcal{\Bar{A} \longrightarrow \mathcal{A}}$ mapping latent actions to the original action space (see Figure \ref{fig:models}).

In our work, we study the case of a discrete action space $\mathcal{A}$, therefore, to train latent model $\psi_e$ and decoder $\delta_d$, it is possible to use the cross-entropy loss\footnote{In continuous action spaces, it is possible to use simply the mean squared error loss between the action and the predicted action using the models.} in Equation (\ref{decoder_loss_discrete}).
\begin{equation}
\begin{split}
\mathcal{L}_\delta(\pmb{\theta}_{\psi_e}, \pmb{\theta}_{\delta_d},\pmb{\theta}_{\psi_e}) &= \mathbb{E}[-\sum_{i=0}^K (a_i\log(\hat{a}_i)) ] \\
&= \mathbb{E}[-\sum_{i=0}^K (a_i\log(\delta_d(\bar{a};\pmb{\theta}_{\delta_d})_i) ] \\
&= \mathbb{E}[-\sum_{i=0}^K (a_i\log(\delta_d(\psi_e(\bar{s},a; \pmb{\theta}_{\psi_e}));\pmb{\theta}_{\delta_d})_i) ]
\end{split}
\label{decoder_loss_discrete}
\end{equation}
where $a_i$ is the $i$-$th$ component of one-hot encoded action $a$ and $\hat{a}_i$ is the $i$-$th$ component of the normalized logit corresponding to the predicted action $\hat{a}$. A similar loss function is employed in \cite{ch2019learning} and \cite{pritz2020joint}.

\subsubsection{The Complete Loss Function}

The total loss function that is minimised for training our neural network models is shown in Equation (\ref{total_loss}) and it is equal to the weighted sum of the four different losses shown in Equation (\ref{transition_loss})-(\ref{contrastive_loss}), and (\ref{decoder_loss_discrete}).

\begin{equation}
\min_{\pmb{\theta}_{\phi_e},\pmb{\theta}_{\psi_e},\pmb{\theta}_{\delta_d},\pmb{\theta}_{\bar{\text{T}}},\pmb{\theta}_{\bar{\text{R}}}} \ \ \ \ \omega_{\Bar{\text{T}}}\mathcal{L}_{\Bar{\text{T}}}(\pmb{\theta}_{\phi_e},\pmb{\theta}_{\bar{\text{T}}},\pmb{\theta}_{\psi_e})  + \omega_{\Bar{\text{R}}}\mathcal{L}_{\Bar{\text{R}}}(\pmb{\theta}_{\phi_e},\pmb{\theta}_{\bar{\text{R}}},\pmb{\theta}_{\psi_e}) +\omega_c\mathcal{L}_c(\pmb{\theta}_{\phi_e},\pmb{\theta}_{\bar{\text{T}}},\pmb{\theta}_{\psi_e}) + \omega_{\delta_d}\mathcal{L}_\delta(\pmb{\theta}_{\psi_e}, \pmb{\theta}_{\delta_d})
\label{total_loss}
\end{equation}

It is worth mentioning that the action encoder $\psi_e$ is affected by the MDP homomorphism losses, in Equation (\ref{total_loss_MDP}), and the action decoder loss, in Equation (\ref{decoder_loss_discrete}). In this way, we aim at learning an action representation that a) exploits symmetries and b) allows reconstruction of the true action space.

\subsection{Optimality of the Policies}

In this section, we first  study the relation between the latent policy $\bar{\pi}: \mathcal{\bar{S}} \longrightarrow \bar{\mathcal{A}}$ and the policy $\pi_o: \mathcal{O} \longrightarrow \mathcal{A}$. Then, we study the relation between the policy $\bar{\pi}$ and the intermediate policy $\pi_i:\mathcal{\bar{S}} \longrightarrow \mathcal{A}$. Eventually, we study the relation between $\pi_o$  and $\pi_i$.


\textbf{\textit{Proposition 1:}} Accordingly to \cite{sutton1999between}, \cite{taylor2008bounding}, \cite{gelada2019deepmdp}, \cite{VanderPol2020}, when the loss in Equation (\ref{total_loss_MDP}) approaches zero, the MDP $\mathcal{\Bar{M}} = \langle \mathcal{\Bar{S}}, \mathcal{\Bar{A}}, \Bar{\text{T}}, \Bar{\text{R}}  \rangle$ is an (approximate) homomorphism of of the original MDP $\mathcal{M} = \langle \mathcal{O}, \mathcal{A}, \text{T}, \text{R}  \rangle$. The optimal latent policy $\bar{\pi}^*: \mathcal{\bar{S}} \longrightarrow \mathcal{A}$ can be lifted to the original MDP by preserving its optimality. Therefore, with reference to Figure \ref{fig:comparison_diagram}, the optimal latent policy $\bar{\pi}^*: \mathcal{\bar{S}} \longrightarrow \bar{\mathcal{A}}$ is equivalent to the optimal policy $\pi_o^*: \mathcal{O} \longrightarrow \mathcal{A}$.





\textbf{\textit{Proposition 2:}} For all deterministic functions $\delta_d$, the gradient  $\nabla_{\pmb{\theta}_{\bar{\pi}}} J_{\bar{\pi}}(\pmb{\theta}_{\bar{\pi}})$ of the performance measure of the latent policy $\bar{\pi}:\mathcal{\bar{S}} \longrightarrow \mathcal{\bar{A}}$ is equivalent to the gradient $\nabla_{\pmb{\theta}_{\bar{\pi}}} J_{\pi_i}(\pmb{\theta}_{\bar{\pi}}, \pmb{\theta}_{\delta_d})$ of the performance measure the intermediate policy  $\pi_i:\mathcal{\bar{S}} \longrightarrow \mathcal{A}$:

\begin{equation}
    \nabla_{\pmb{\theta}_{\bar{\pi}}} J_{\pi_i}(\pmb{\theta}_{\bar{\pi}}, \pmb{\theta}_{\delta_d}) = \nabla_{\pmb{\theta}_{\bar{\pi}}} J_{\bar{\pi}}(\pmb{\theta}_{\bar{\pi}}) 
\end{equation}

Therefore ascending the gradient of $\bar{\pi}$ is equivalent to ascending the gradient of $\pi_i$. The complete proof is shown in Appendix \ref{appendixA}.

\textbf{\textit{Proposition 3:}} As consequence of \textit{Proposition 1} and \textit{Proposition 2},  an optimal internal policy $\pi_i^*: \bar{\mathcal{S}} \longrightarrow \mathcal{A}$ is equivalent to an optimal policy $\pi_o^*$:  $\mathcal{O} \longrightarrow \mathcal{A}$. Thus, an optimal latent policy $\bar{\pi}^*$ for the MDP $\mathcal{\bar{M}}$ is equivalent to an optimal intermediate policy $\pi_i^*$ and to an optimal policy $\pi_o^*$ for the original MDP $\mathcal{M}$.




\subsection{Neural Network Architectures}

\begin{figure*}[h!]
\centering
    \begin{subfigure}{0.49\textwidth}
    \centering
\includegraphics[width=0.85\linewidth,page=9]{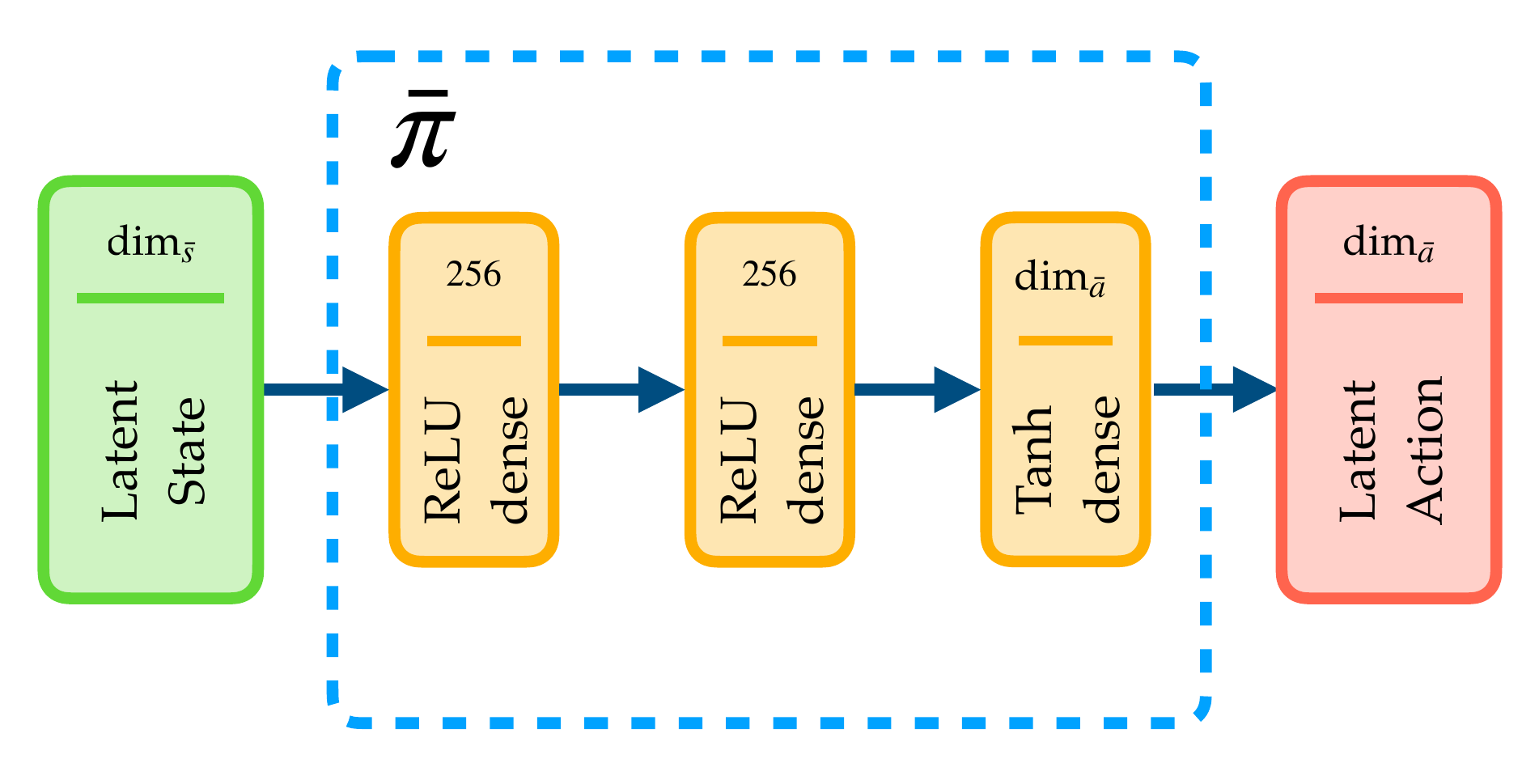}
\captionsetup{justification=centering}
    \caption{Actor Model}
\label{fig:actor}
    \end{subfigure} \hfill
    \begin{subfigure}{0.49\textwidth}
    \centering
\includegraphics[width=0.85\linewidth,page=10]{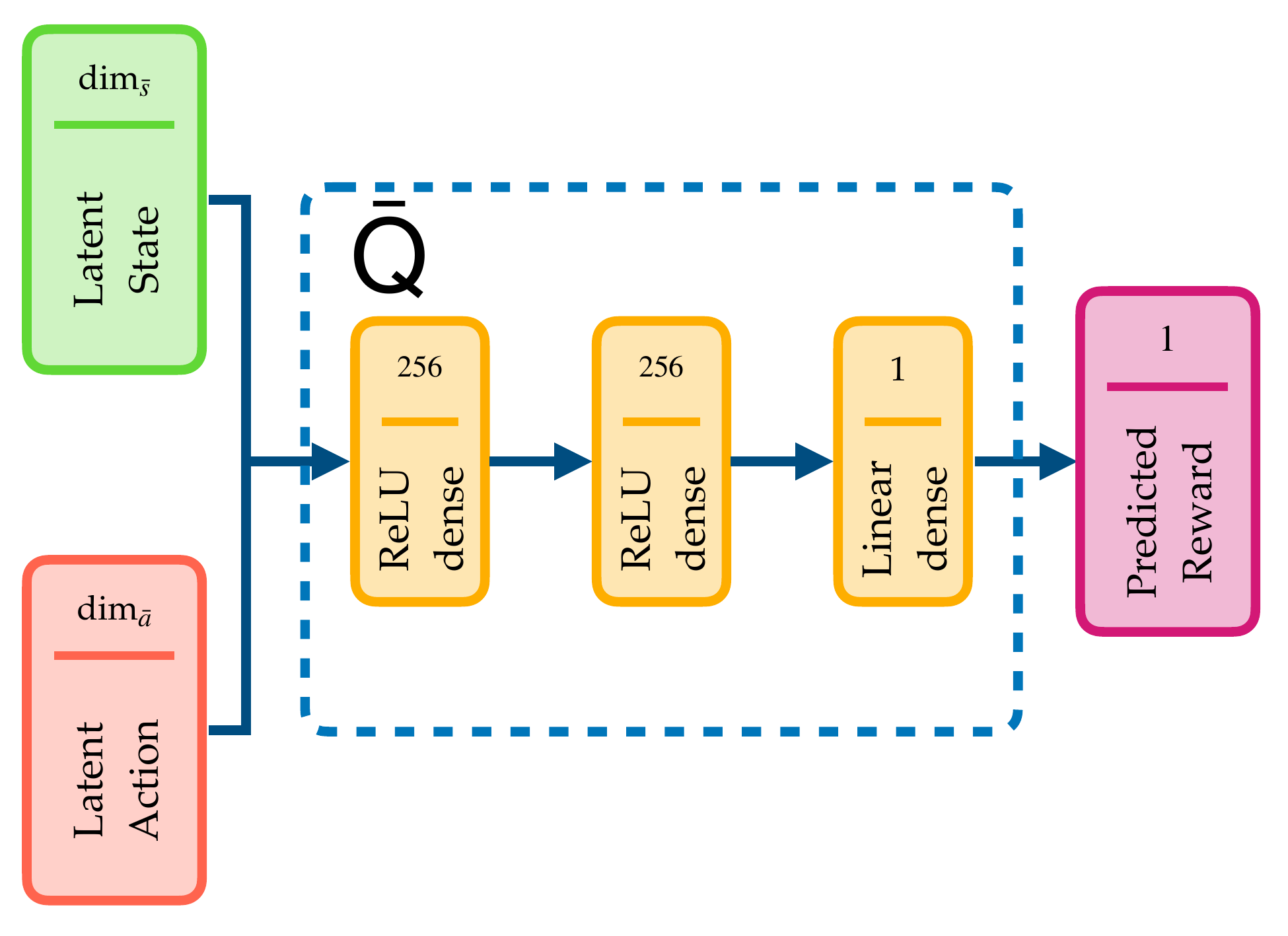}
\captionsetup{justification=centering}
    \caption{Critic Model}
\label{fig:critic}
    \end{subfigure} \hfill
\caption{Neural network architecture of the latent policy $\bar{\pi}$ and the latent action-value function $\bar{\text{Q}}$}
    \label{fig:TD3NNs}
    \end{figure*}

For learning the latent policy $\bar{\pi}$, any Reinforcement Learning algorithm that can deal with continuous state and action space can be used. Here, we use TD3 (see Section \ref{subsec:TD3}) with the implementation provided in \cite{fujimoto2018addressing}. Actor and critic networks are composed of two fully connected layers with 256 units with ReLU activation. The output layer of the actor has Tanh activation and outputs latent actions, while the critics have linear activation and output the Q-values of the state-action pairs. The actor and a critic neural networks are shown in Figure \ref{fig:TD3NNs}.

\begin{figure}[h!]
    \centering
    \includegraphics[width=0.85\linewidth, page=4]{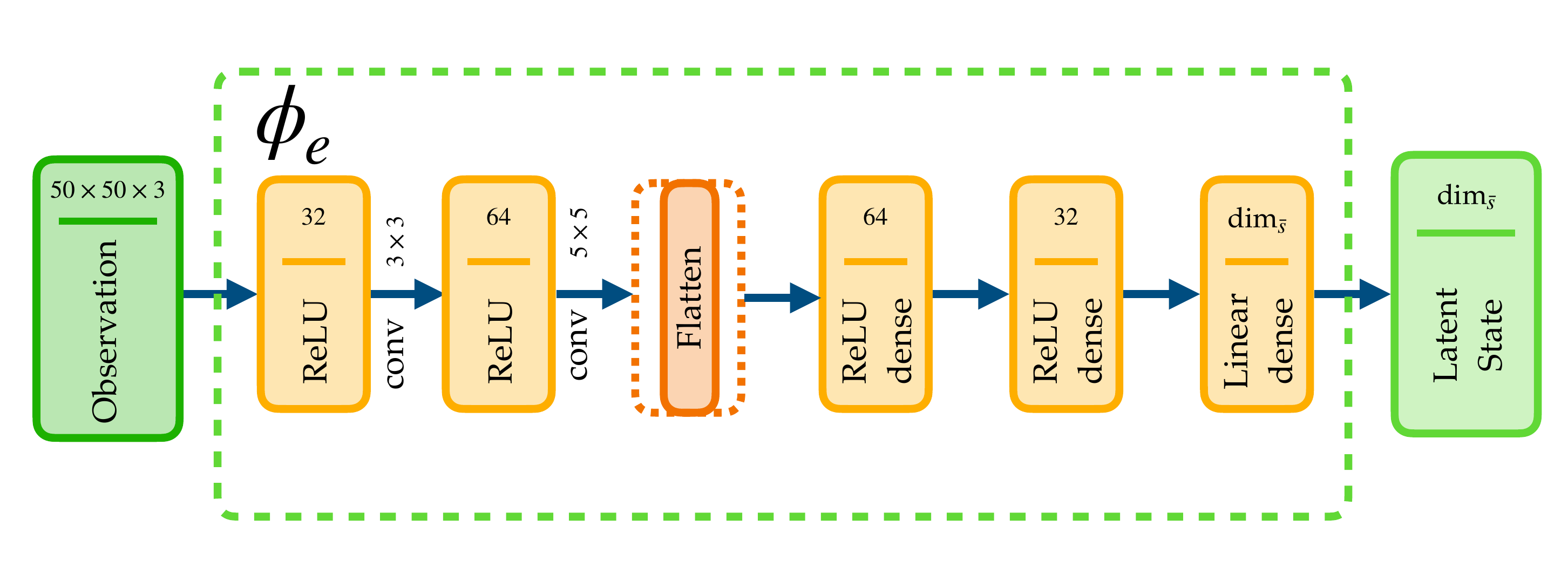}
    \caption{Neural network architecture of the observation encoder $\phi_e$}
    \label{fig:encoderNN}
\end{figure}

\begin{figure*}[h!]
\centering
    \begin{subfigure}{0.51\textwidth}
    \centering
\includegraphics[width=1.2\linewidth,page=5]{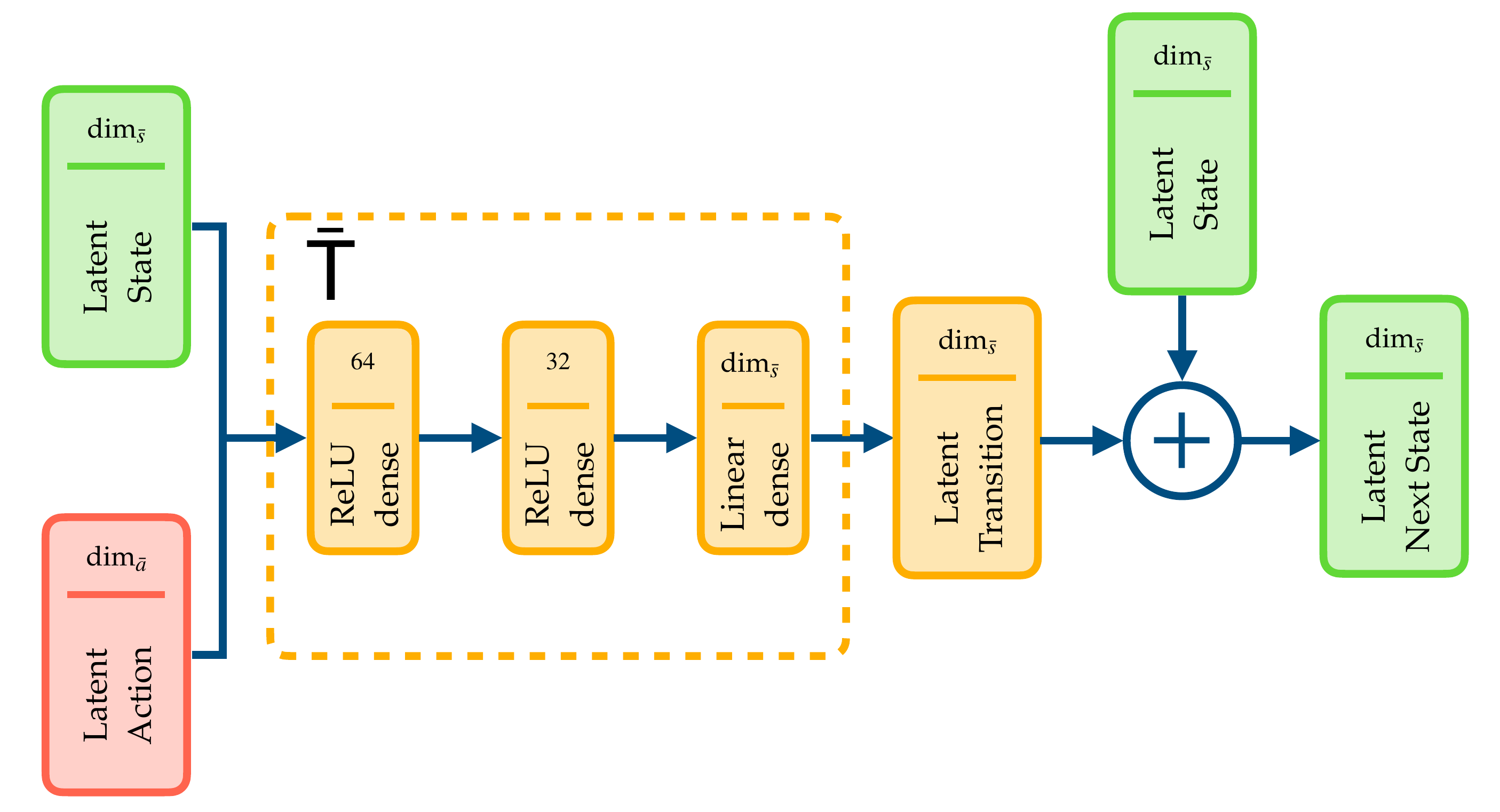}
\captionsetup{justification=centering}
    \caption{Transition Model}
\label{fig:tmodel}
    \end{subfigure} 
    \begin{subfigure}{0.51\textwidth}
    \centering
\includegraphics[width=0.85\linewidth,page=6]{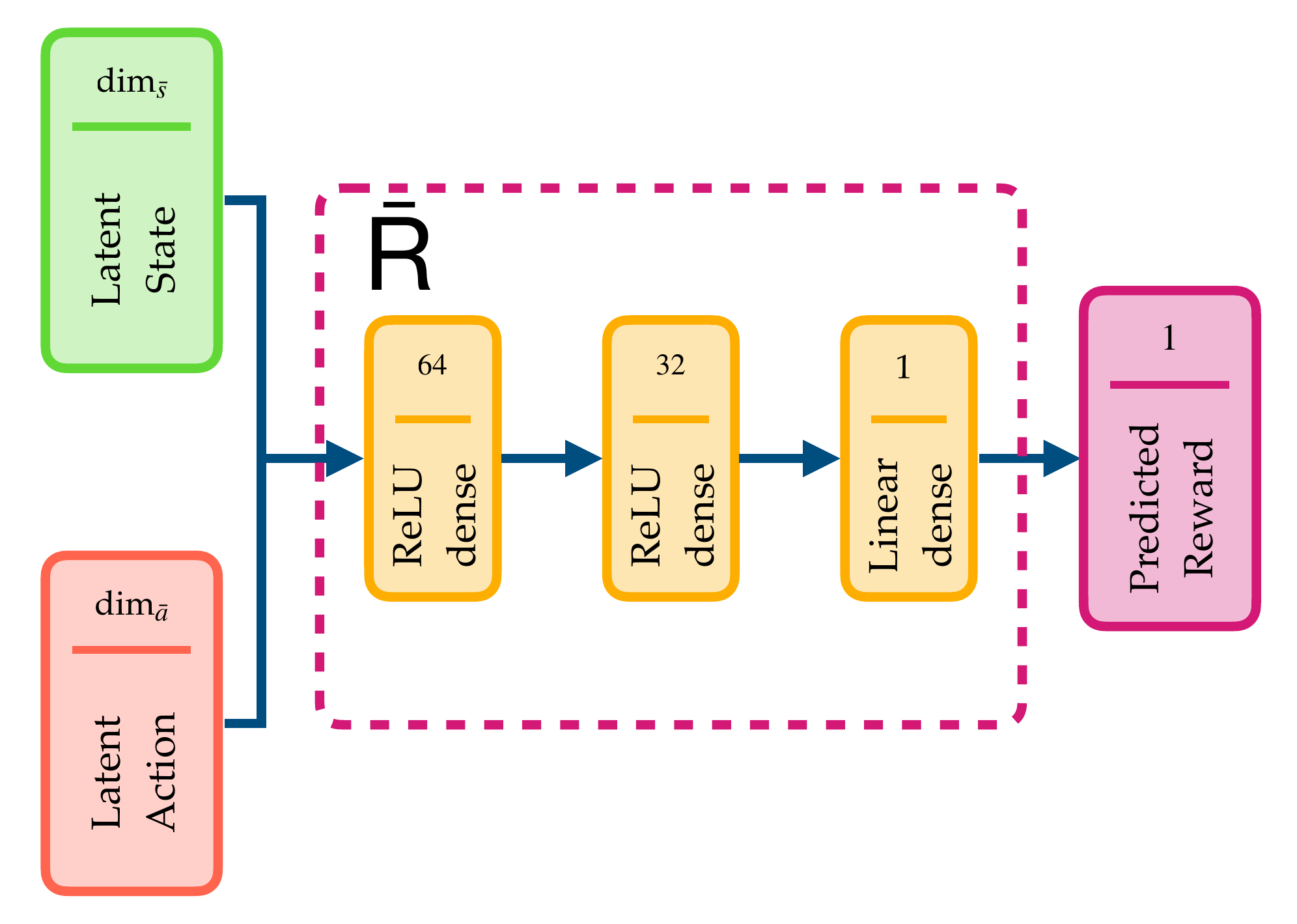}
\captionsetup{justification=centering}
    \caption{Reward Model}
\label{fig:rmodel}
    \end{subfigure} 
\caption{Neural network architecture of the transition model $\bar{\text{T}}$ and reward model $\bar{\text{R}}$.}
    \label{fig:transitionrewardNN}
    \end{figure*}

The encoder $\phi_e$ is composed of two convolutional layers, with 32 and 64 filters of size $3 \times 3$ and $5 \times 5$ respectively,  with ReLU activations, two fully-connected layers, with 64 and 32 units, with ReLU activation, and a final fully-connected layer with linear activation outputting latent states. The complete architecture can be seen in Figure \ref{fig:encoderNN}. 
The transition model and the reward model share a similar architecture with two fully-connected layers, with 64 and 32 units, and ReLU activation respectively, and an output layer with linear activation, as shown in Figure \ref{fig:transitionrewardNN}. 
Similar architectures are employed in \cite{VanderPol2020}.     
   
Eventually, the action encoder comprises two fully connected layers, with 64 and 32 units and ReLU activation. The output layer has tanh activation to bound the latent action space in $[-1,1]$. The action decoder has a similar architecture except a softmax output activation to map latent actions to one-hot encoded action of the original action space. A similar architecture is employed in \cite{ch2019learning}.
\begin{figure*}[h!]
\centering
    \begin{subfigure}{0.49\textwidth}
    \centering
\includegraphics[width=0.85\linewidth,page=7]{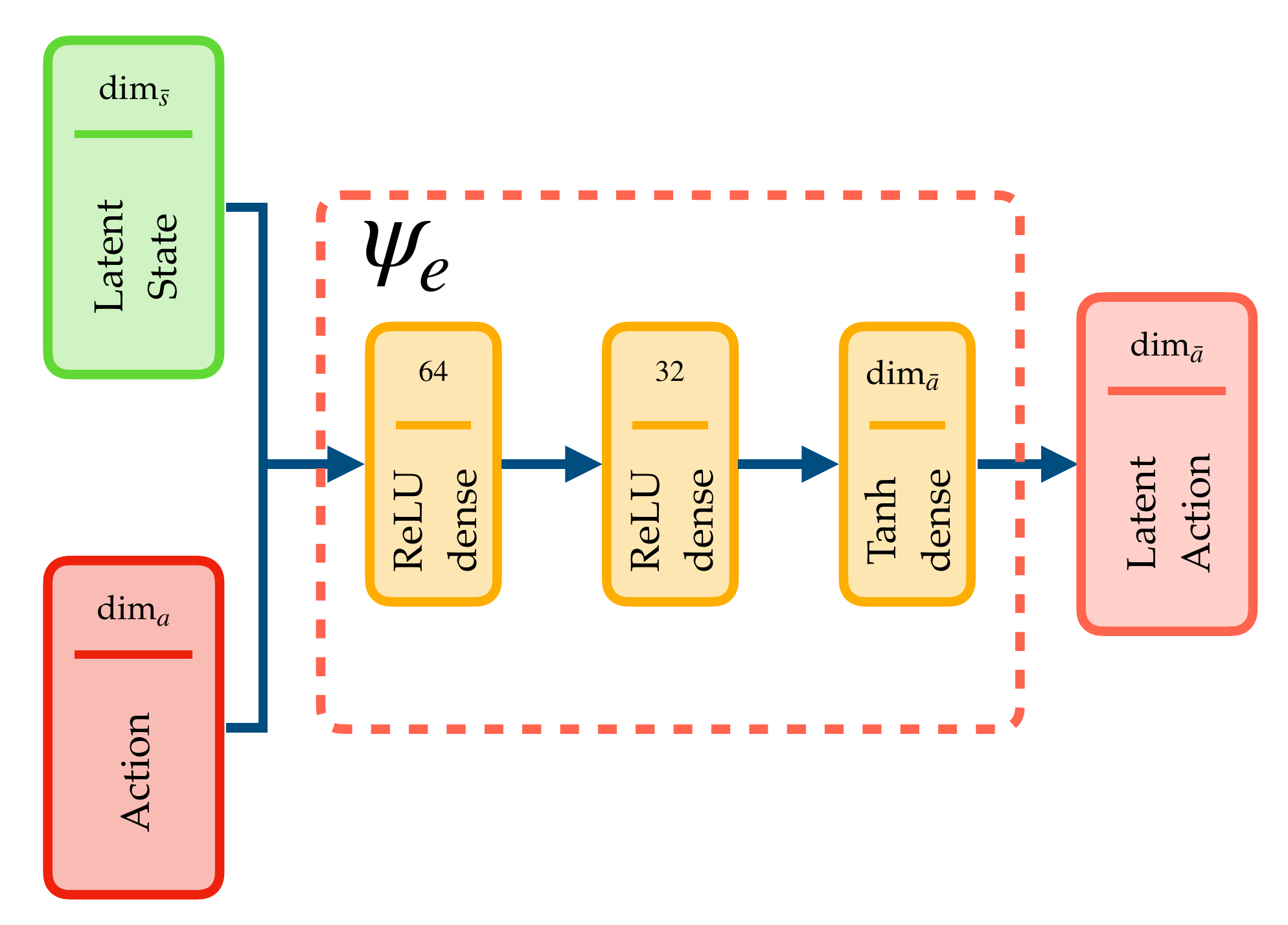}
\captionsetup{justification=centering}
    \caption{Action Encoder Model}
\label{fig:aenc}
    \end{subfigure} \hfill
    \begin{subfigure}{0.49\textwidth}
    \centering
\includegraphics[width=0.85\linewidth,page=8]{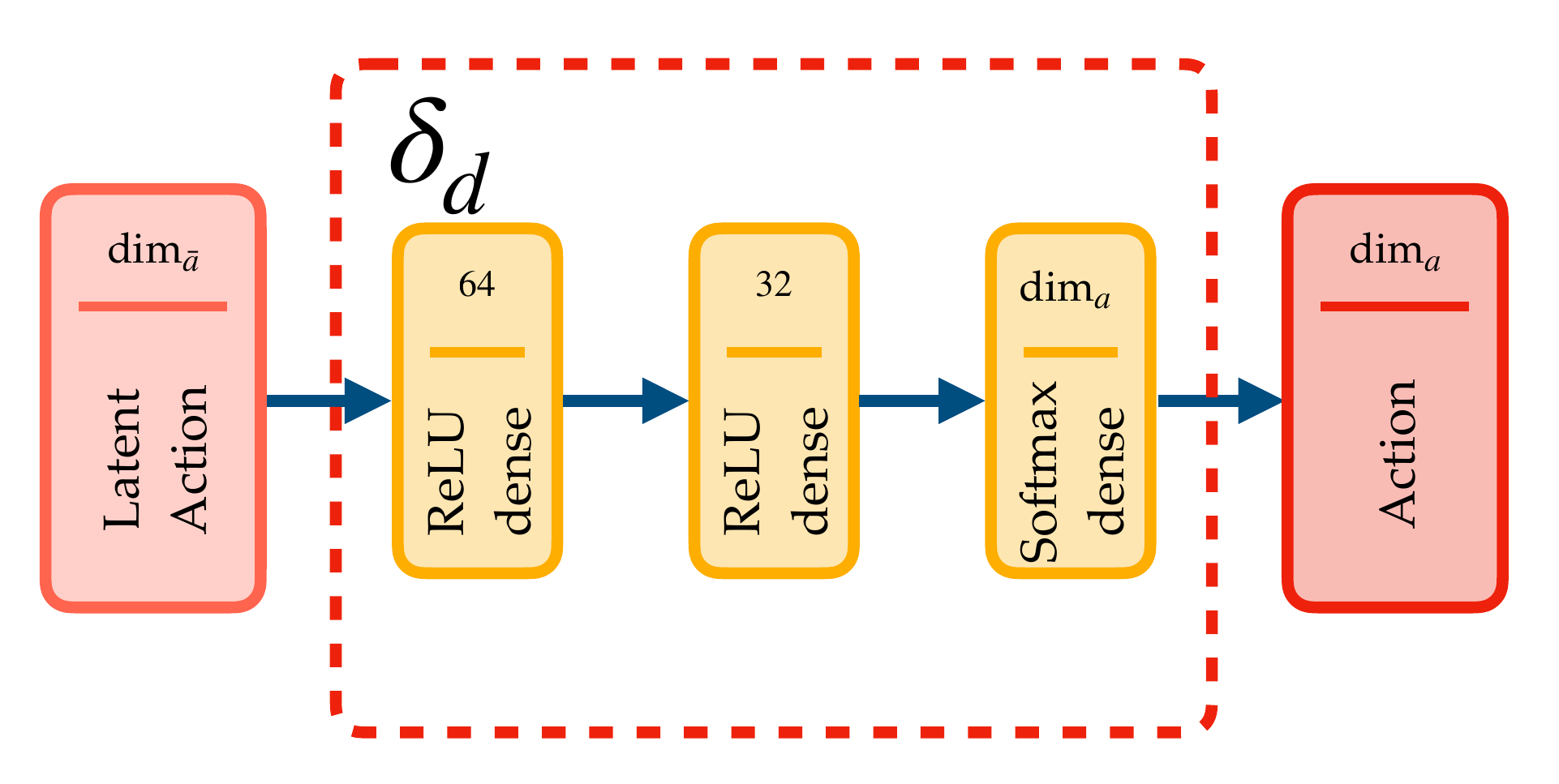}
\captionsetup{justification=centering}
    \caption{Action Decoder Model}
\label{fig:adec}
    \end{subfigure} \hfill
\caption{Neural network architecture of the action encoder model $\psi_e$ and the action decoder model $\delta_d$.}
    \label{fig:actionencdecNN}
    \end{figure*}

\section{Experimental Design}\label{sec:ExperimentalDesign}
\subsection{Grid-World}
\begin{figure*}[ht!]
\centering
    \begin{subfigure}{0.3\textwidth}
    \centering
\includegraphics[width=0.75\linewidth]{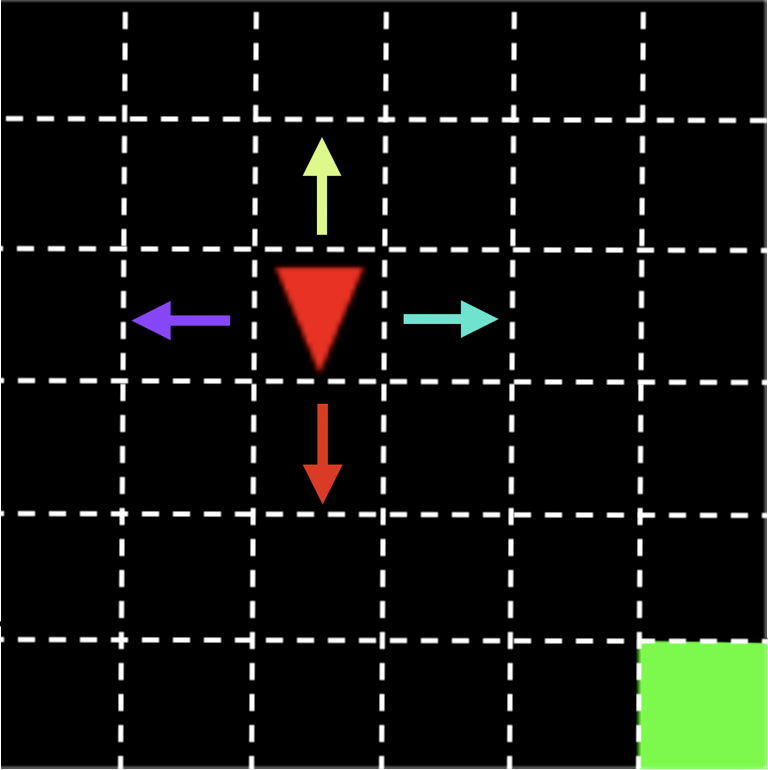}
\captionsetup{justification=centering}
    \caption{$6\times6$ maze - 1 object}
\label{fig:env-1obj}
    \end{subfigure} \hfill
    \begin{subfigure}{0.3\textwidth}
    \centering
\includegraphics[width=0.75\linewidth]{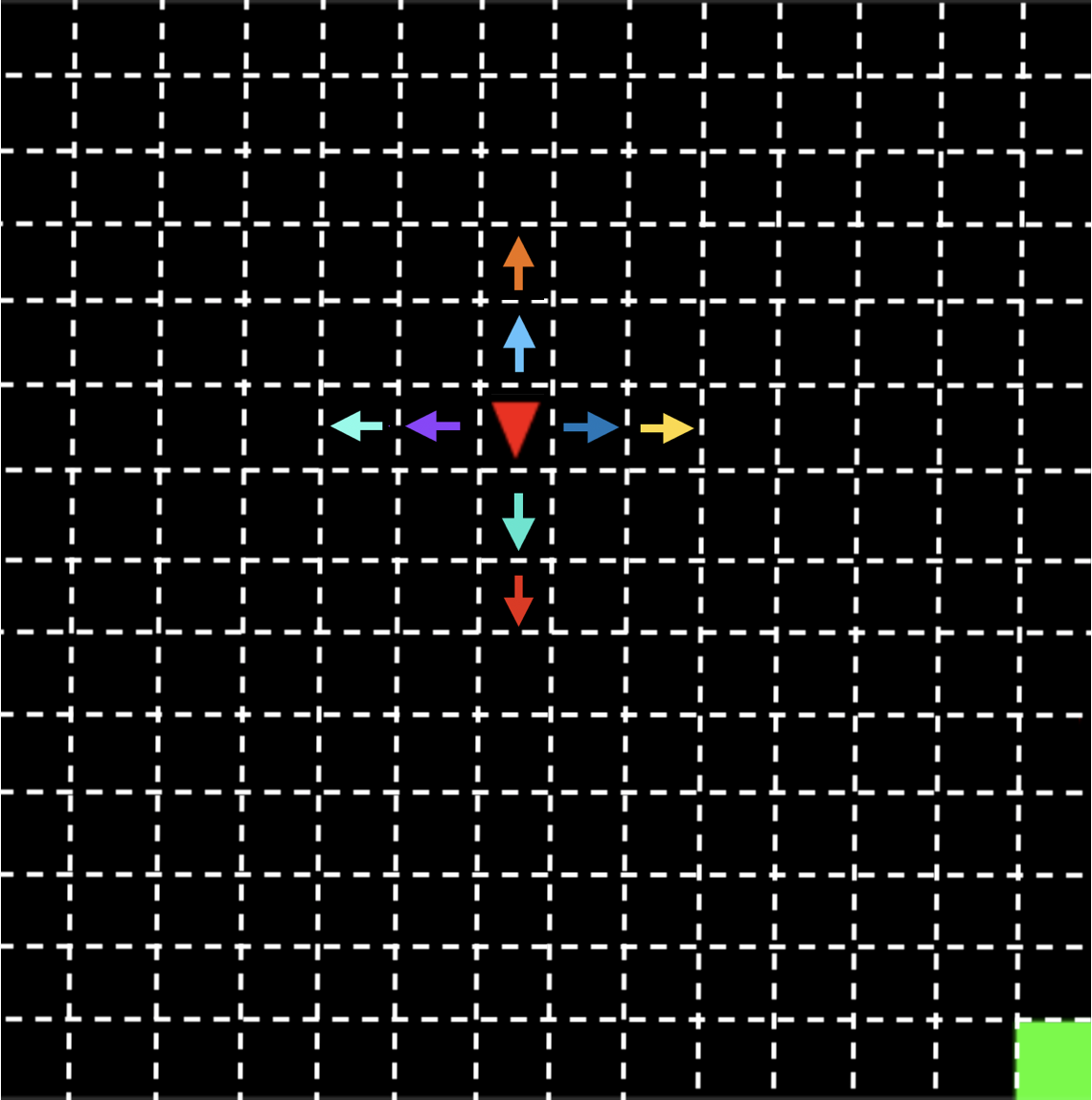}
\captionsetup{justification=centering}
    \caption{$14\times14$ maze - 1 object}
\label{fig:env-2obj}
    \end{subfigure} \hfill
\begin{subfigure}{0.3\textwidth}
\centering
\includegraphics[width=0.75\linewidth]{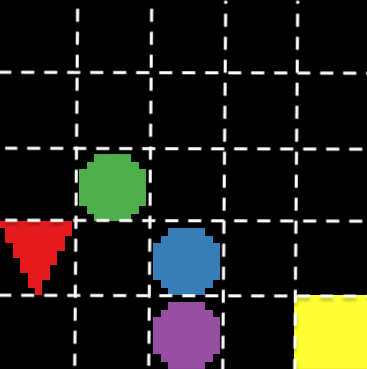}
\captionsetup{justification=centering}
    \caption{$5\times5$ maze - 1 objects, 3 distractors}
\label{fig:env-3obj}
    \end{subfigure} 
\caption{Observations from different grid-world environments. The triangles indicate the robots controlled by the agent, the squares the target positions, the circles are visual distractors, and the arrows represent the different actions of the agent. }
    \label{fig:grid_envs}
    \end{figure*}

The grid-world can be seen as the simple mobile robot navigation problem, where the agent has to navigate the robot to a target cell of the grid, and it is enforced to move along the underlying grid. At each training episode, the robot is randomly spawned in different positions of the maze. Grid-worlds are used in \cite{VanderPol2020}, \cite{kipf2019contrastive}, \cite{ch2019learning}, and \cite{pritz2020joint}. In our experiments, the agent can observe the maze through RGB images of size $50 \times 50$. Examples of mazes and agent's observations can be found in Figure \ref{fig:grid_envs}. For the grid-world experiments, we adapted the environments in \cite{gym_minigrid} and \cite{kipf2019contrastive}. 

We use for all the experiments a distance-based reward function, as shown in Equation (\ref{reward_function_grid}).

\begin{equation}
    \small
    \text{R}(s, a)=\begin{cases}
    r_{\text{reached}}, & s=s_{\text{goal}},\\
    -\eta d, & \text{otherwise}.
    \end{cases}
    \label{reward_function_grid}
\end{equation}
where $r_{\text{reached}}$ is a bonus for reaching the goal position, $d$ is the Manhattan distance robot-goal normalized over the number of cells of the maze, and $\eta$ is a scaling factor. Distance-based reward functions are a natural choice for robot navigation tasks.

We experiment in:
\begin{itemize}
    \item $6  \times 6$ maze in which the agent has to steer a single robot (red triangle in Figure \ref{fig:env-1obj}) to a target position (green square) by choosing among four possible actions.
    \item $14  \times 14$ maze in which the agent has to steer a single robot (red triangle in Figure \ref{fig:env-2obj}) to a target position (green square) by choosing among eight possible actions.
    \item $5  \times 5$ maze in which the agent has to steer a single robot (red triangle in Figure \ref{fig:env-3obj}) to a target position (yellow square) by choosing among four possible actions per object. During the training of the policy, up to three unseen distractors (circles) randomly move across the maze. 
    
\end{itemize}

\subsection{Mobile Robot Navigation}

Secondly, we test our approach on a simple mobile robot navigation task with continuous underlying state space. The mobile robot (Pioneer p3dx) is simulated on VRep \cite{rohmer2013v} using the PyRep interface \cite{james2019pyrep}. The agent receives $48 \times 48$  RGB images coming from an onboard camera, and its action space is composed of three and eight different discrete actions. 

\begin{figure*}[ht!]
\centering
    \begin{subfigure}{0.3\textwidth}
    \centering
\includegraphics[width=0.75\linewidth]{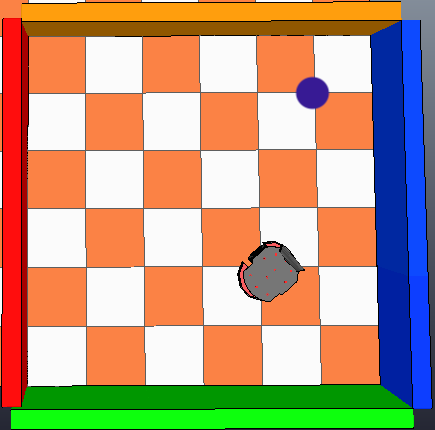}
\captionsetup{justification=centering}
    \caption{Robot navigation task}
\label{fig:env-robot}
    \end{subfigure} 
    \begin{subfigure}{0.3\textwidth}
    \centering
\includegraphics[width=0.75\linewidth]{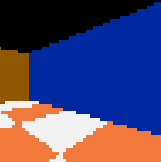}
\captionsetup{justification=centering}
    \caption{Example of observation}
\label{fig:env-obsrobot}
    \end{subfigure}
\caption{The agent needs to steer the robot to the target (purple circle in Figure \ref{fig:env-robot}) by relying on RGB images coming from an onboard camera (Figure \ref{fig:env-obsrobot}).}
    \label{fig:robot_envs}
    \end{figure*}
    
The reward function is similar to the one used in the grid-world but with a penalty for colliding with the walls. The complete reward function is shown in Equation (\ref{reward_function_robot}).

\begin{equation}
    \small
    \text{R}(s, a)=\begin{cases}
    r_{\text{reached}}, & d \leq d_{min},\\
    r_{\text{crashed}}, & s=s_{ts}, \\
    -\eta d, & \text{otherwise}.
    \end{cases}
    \label{reward_function_robot}
\end{equation}
where $r_{\text{reached}}$ is a bonus for reaching the goal position, $r_{\text{crashed}}$ is a penalty for colliding with an obstacle, i.e. reaching a terminal state $s_{ts}$, $d$ is the Euclidean distance robot-goal, and $\eta$ is a scaling factor.

\subsection{Comparison of the Learned State Representations}\label{subsec:compare_state_rep}

To assess the validity of our approach, we qualitatively compare the learned state representation of our method with:
\begin{itemize}
    \item MDP-H: adaptation of the plannable MDP homomorphism framework proposed in \cite{VanderPol2020} in which we learn the state representation by means of the auxiliary losses in Equation (\ref{transition_loss}), (\ref{reward_loss}) and (\ref{contrastive_loss}), but without the action representation module $\psi_e$ and only using the true action $a$.
    \item D-MDP: adaptation of the Deep MDP framework proposed in \cite{gelada2019deepmdp} in which we learn the state representation by means of the auxiliary losses in Equation (\ref{transition_loss}) and (\ref{reward_loss}), but without the action representation module $\psi_e$ and only using the true action $a$.
     \item JSAE: adaptation of the joint state-action embeddings framework proposed in \cite{pritz2020joint}.
    \item JSAE-C: Adaptation of the joint state-action embeddings framework proposed in \cite{pritz2020joint} with the addition of the contrastive loss, in Equation (\ref{contrastive_loss}), for preventing the trivial embedding in which all states are mapped to the zero vector.
\end{itemize}

For the fairness of comparison, we train all the neural networks using the same data-set of samples collected through random interaction with the environments, the same network architectures for state encoder $\phi_e$, latent transition model $\bar{\text{T}}$, and reward model $\bar{\text{R}}$, same learning rate, batch size, latent state space dimensionality, and three random seeds. 

The list of hyperparameters used is shown in Table \ref{ch7:tab_1}.
\begin{table} [h!]
\centering
\scalebox{0.9}{
\begin{tabular}{ ||c|c|| } 
\hline 
Hyperparameter & Value \\
\hline \hline
Latent state dimension ($\text{dim}_{\bar{s}}$)  & 10  \\ 
Latent transition dimension ($\text{dim}_{\bar{s}}$)  & 10  \\ 
Latent action dimension ($\text{dim}_{\bar{a}}$) & 5  \\ 
Learning rate & 0.0005   \\
Batch size & 256  \\
Training Epochs & 100 \\
optimiser & ADAM \\
\hline
\end{tabular}}
\caption{Hyperparameters of the  experiments}
\label{ch7:tab_1}
\end{table}

\subsection{Comparison of the Learned Policies}

After learning the state (and action) representation, we aim at learning the optimal policy given such a representation. We, therefore, compare, in terms of the average number of steps the agents take over training, the performance of our approach, learning a continuous latent policy given a fixed state and action representation, with the performance of a Deep Q-Network agent \cite{mnih2015human}, or  DQN, mapping latent states directly to actions. In the latter, the state representation is learned with the method proposed in \cite{VanderPol2020}, but without the discretisation step employed by the authors. The discretisation of the latent state space would limit the applicability only to MDPs with underlying discrete state space. While this is true for the grid-worlds in Figure \ref{fig:grid_envs}, in the case of mobile robot navigation, in Figure \ref{fig:robot_envs}, the underlying state space is continuous. 

Similarly to \cite{jaderberg2016reinforcement}, we are interested in the best-performing agents; therefore, we train each policy (TD3 and DQN) using ten different seeds, but we plot the mean and the variance of the best three seeds per algorithm. 

The list of hyperparameters used is shown in Table \ref{ch7:tab_2}.
\begin{table} [h!]
\centering
\scalebox{0.9}{
\begin{tabular}{ ||c|c|| } 
\hline 
Hyperparameter & Value \\
\hline \hline
Latent state dimension ($\text{dim}_{\bar{s}}$)  & 10  \\ 
Latent action dimension ($\text{dim}_{\bar{a}}$) TD3 & 5  \\ 
Action dimension DQN & 3,4,8 \\
Learning rate DQN & 0.0005   \\
Learning rate Actor & 0.0005\\
Learning rate Critic & 0.0005\\
Batch size & 64 \\
optimiser & ADAM \\
$\epsilon$-greedy coefficient & 0.25 \\
Random noise $\sigma$  & 0.35 \\
\hline
\end{tabular}}
\caption{Hyperparameters of the  experiments}
\label{ch7:tab_2}
\end{table}




\section{Results}\label{sec:Results}

\subsection{Grid-World}
\subsubsection{Comparison of the  Learned Representations}

\begin{figure*}[ht!]
\centering
    \begin{subfigure}{0.33\textwidth}
\includegraphics[width=1.0\linewidth,page=1]{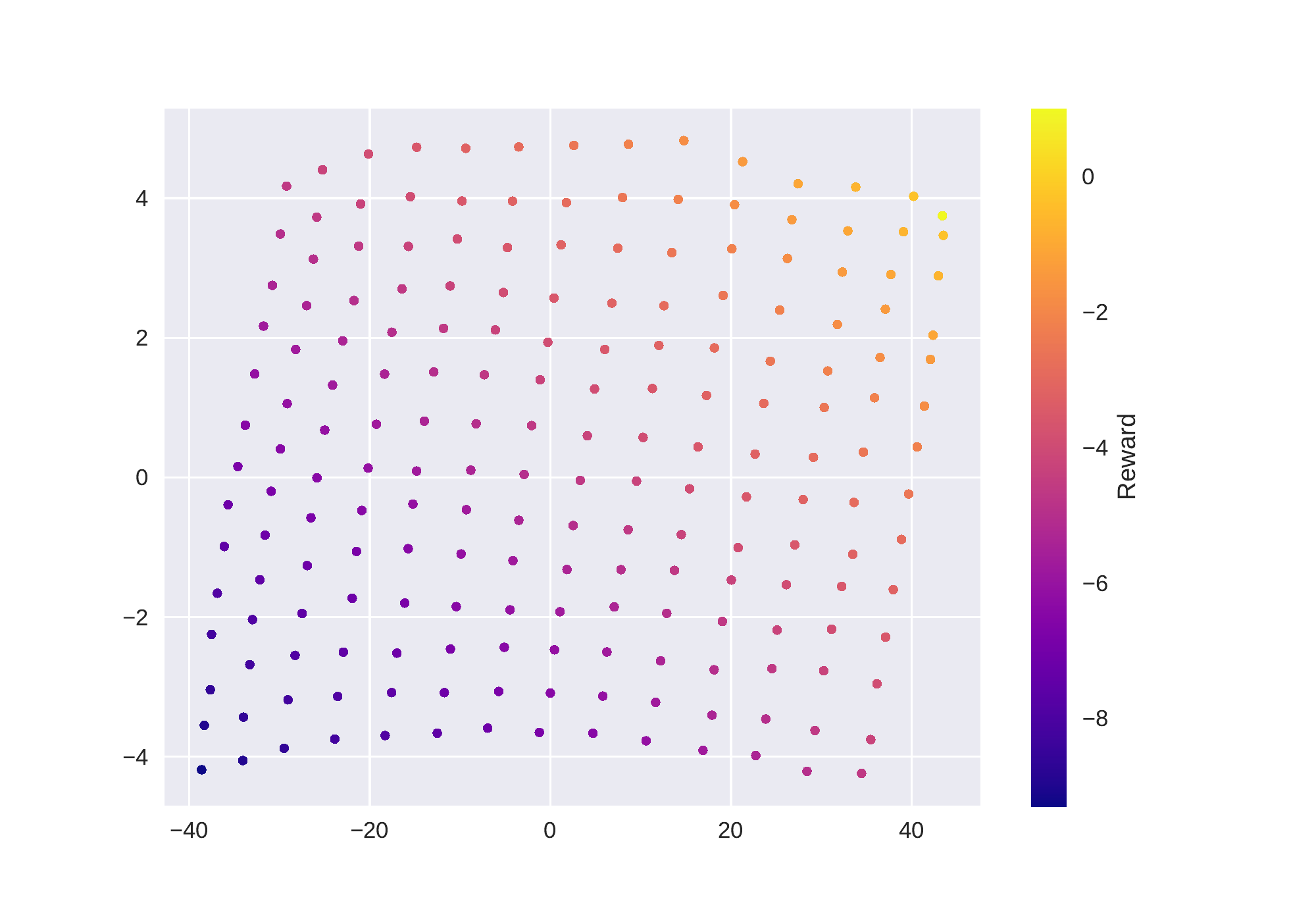}
\captionsetup{justification=centering}
    \caption{Our}
\label{fig:our14}
    \end{subfigure} \hfill
    \begin{subfigure}{0.33\textwidth}
\includegraphics[width=1.0\linewidth,page=1]{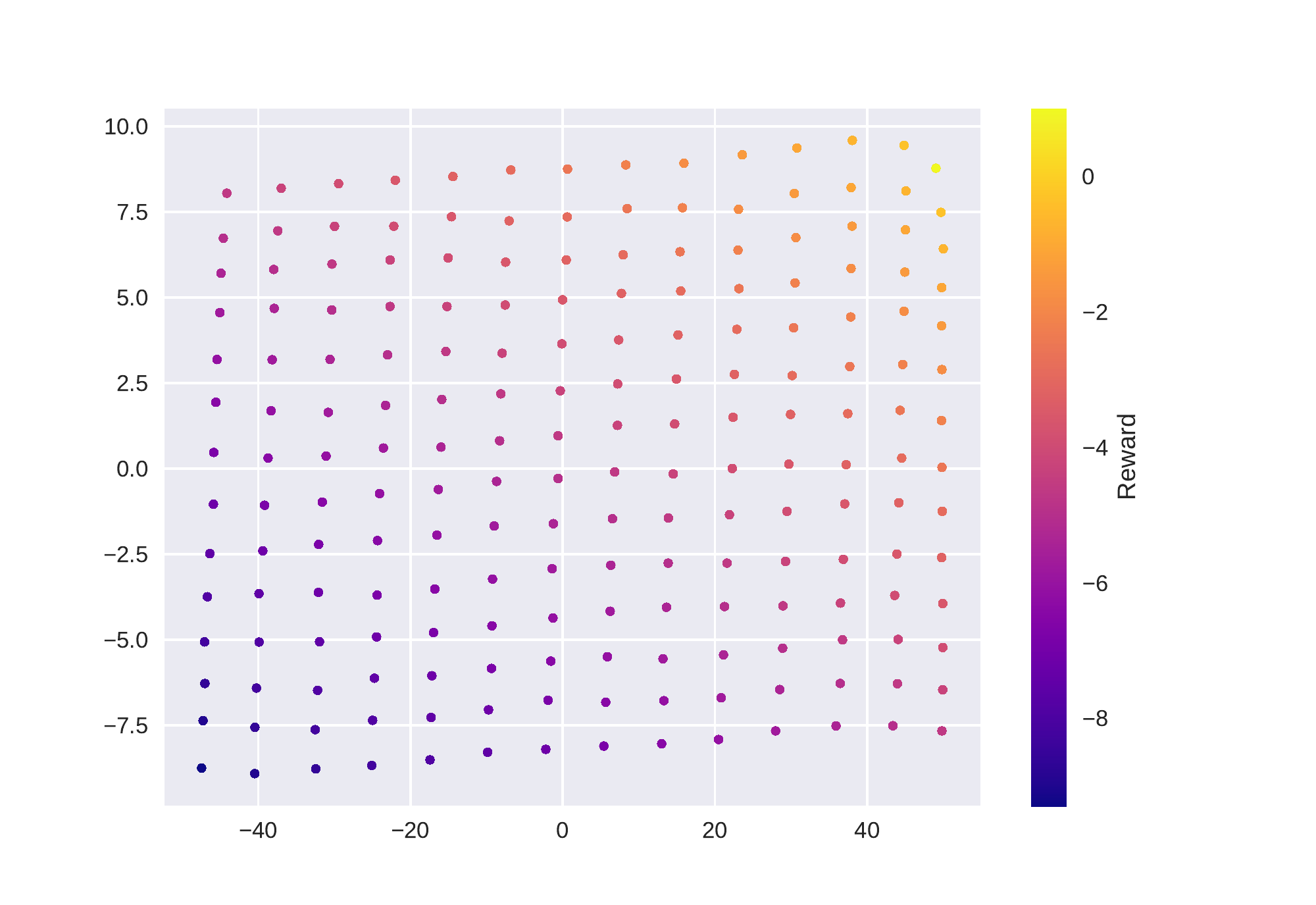}
\captionsetup{justification=centering}
    \caption{MDP-H \cite{VanderPol2020}}
\label{fig:planmdp14}
    \end{subfigure} \hfill
\begin{subfigure}{0.33\textwidth}
\includegraphics[width=1.0\linewidth,page=1]{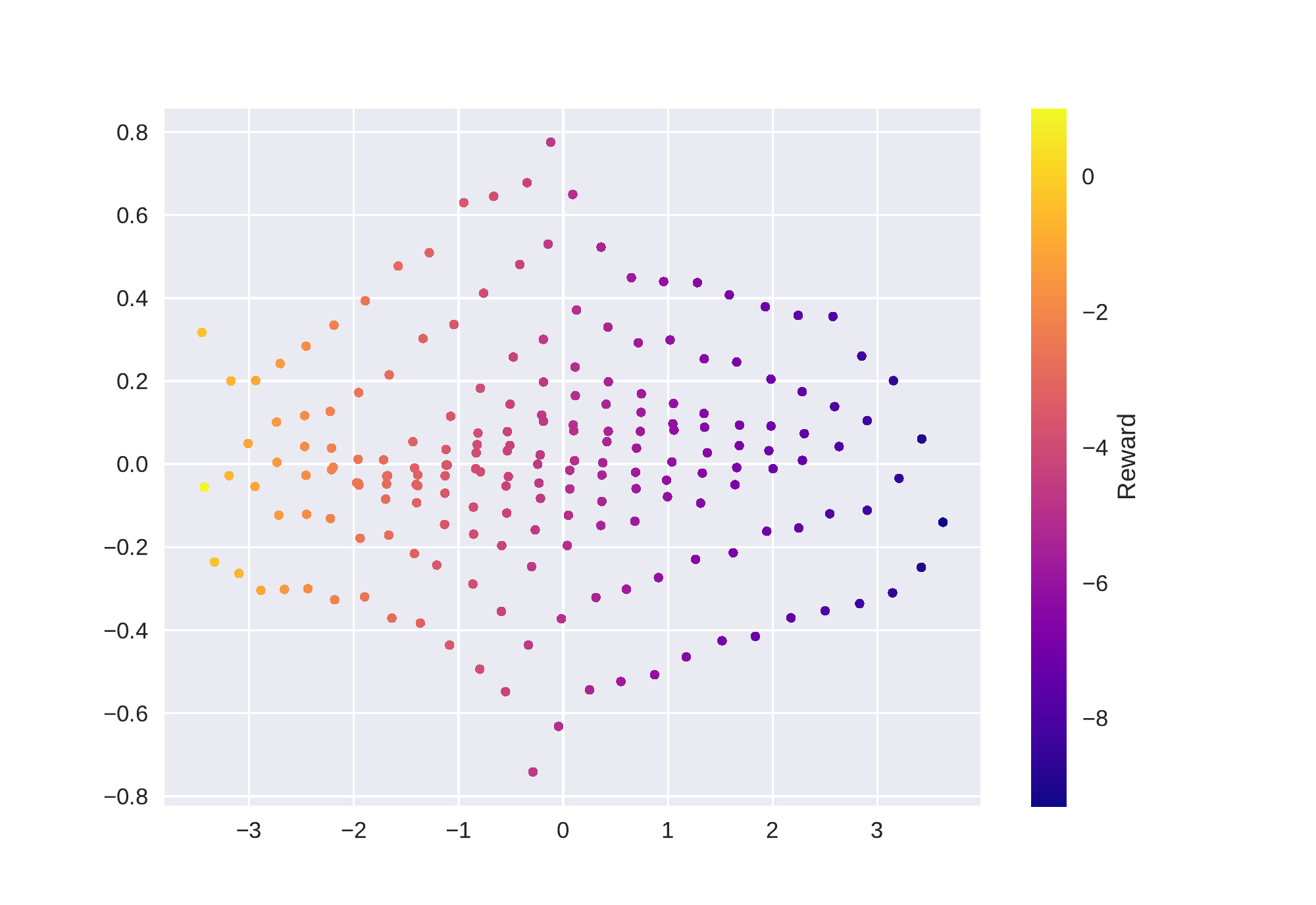}
\captionsetup{justification=centering}
    \caption{D-MDP \cite{gelada2019deepmdp}}
\label{fig:deepmdp14}
    \end{subfigure} \hfill
    \begin{subfigure}{0.33\textwidth}
\includegraphics[width=1.0\linewidth,page=1]{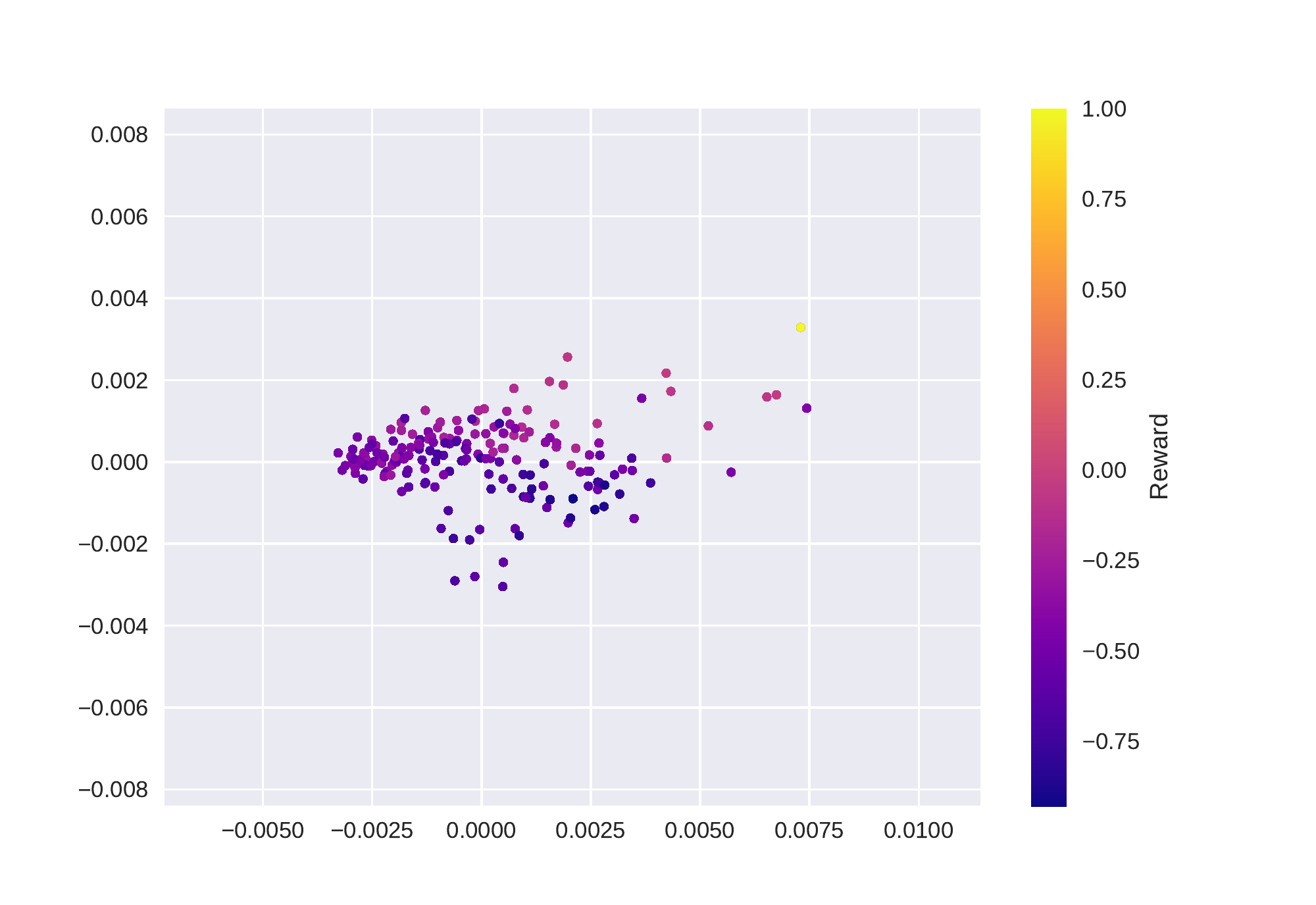}
\captionsetup{justification=centering}
    \caption{JSAE \cite{pritz2020joint}}
\label{fig:jsae14}
    \end{subfigure}
        \begin{subfigure}{0.33\textwidth}
\includegraphics[width=1.0\linewidth,page=1]{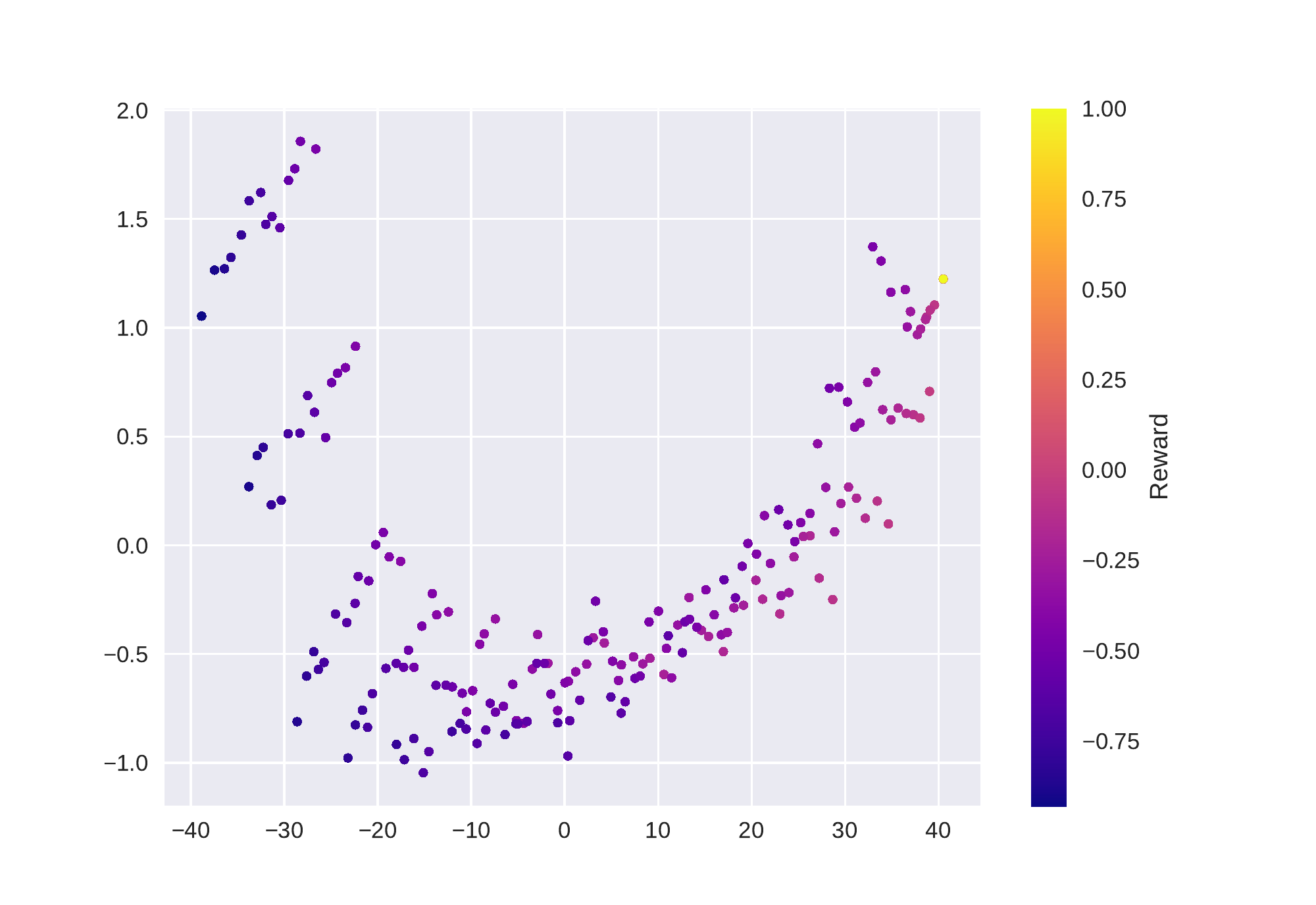}
\captionsetup{justification=centering}
    \caption{JSAE-C \cite{pritz2020joint}}
\label{fig:jsaec14}
    \end{subfigure}
\caption{First two principal components of learned state representations of the $14\times14$ grid-world in Figure \ref{fig:env-2obj}.The color of the state predictions is done accordingly to the reward function in Equation (\ref{reward_function_grid}).}
    \label{fig:state_representation_14x14maze}
    \end{figure*}

We first analyse the learned state representations obtained with the different approaches discussed in Section \ref{subsec:compare_state_rep} by plotting the state predictions generated by encoding a set of randomly collected observations of the different mazes. The state representation learned in the $14\times14$ grid-world are shown in Figure \ref{fig:state_representation_14x14maze}. Additionally, the learned state representation in the $6\times6$ grid-world are shown in Appendix B (Figure \ref{fig:state_representation_6x6maze}). 
Only our approach and the plannable MDP homomorphism framework \cite{VanderPol2020} can retrieve the underlying grid structure of the true state space. The Deep MDP \cite{gelada2019deepmdp} can still retrieve a partial structure, while the JSAE struggles even with the addition of the contrastive loss. 

\begin{figure*}[ht!]
\centering
    \begin{subfigure}{0.33\textwidth}
    \centering
\includegraphics[width=1.0\linewidth,page=8]{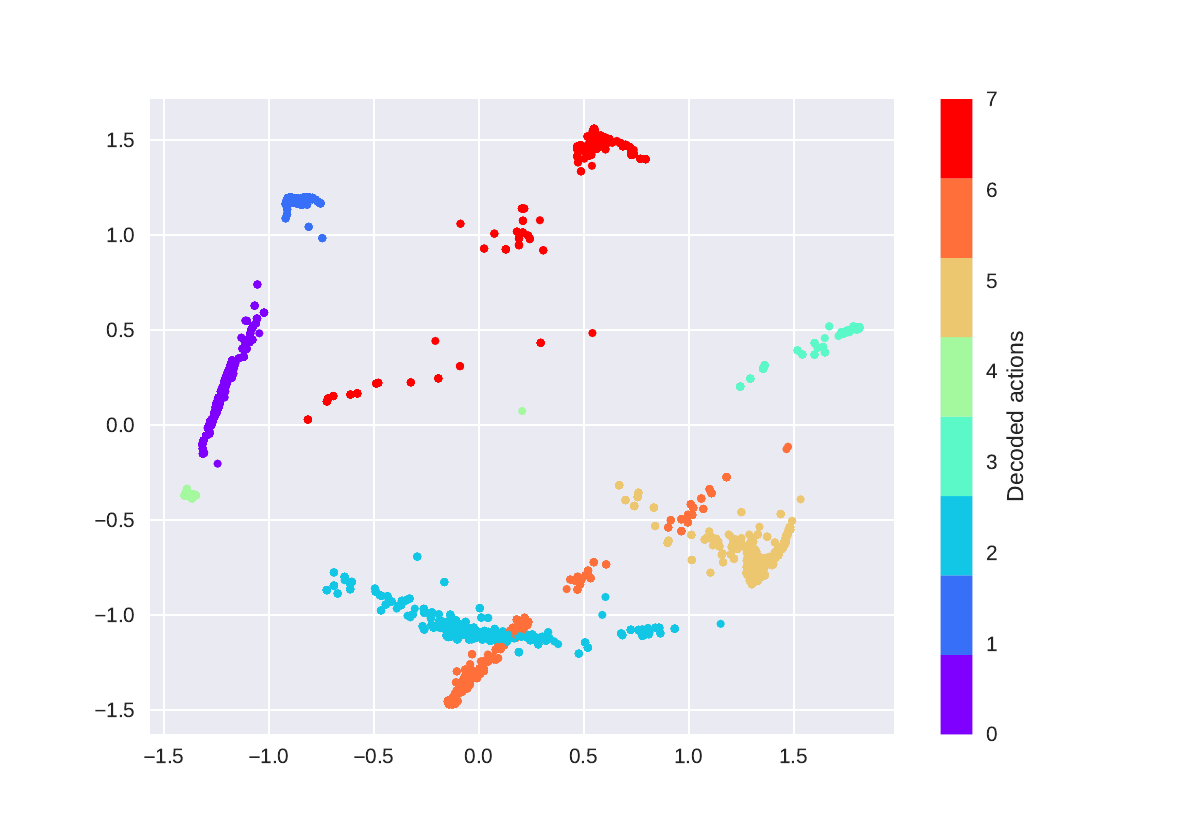}
\captionsetup{justification=centering}
    \caption{Maze $14\times14$, latent actions}
\label{fig:maze14a}
    \end{subfigure} 
\begin{subfigure}{0.33\textwidth}
\centering
\includegraphics[width=1.0\linewidth,page=11]{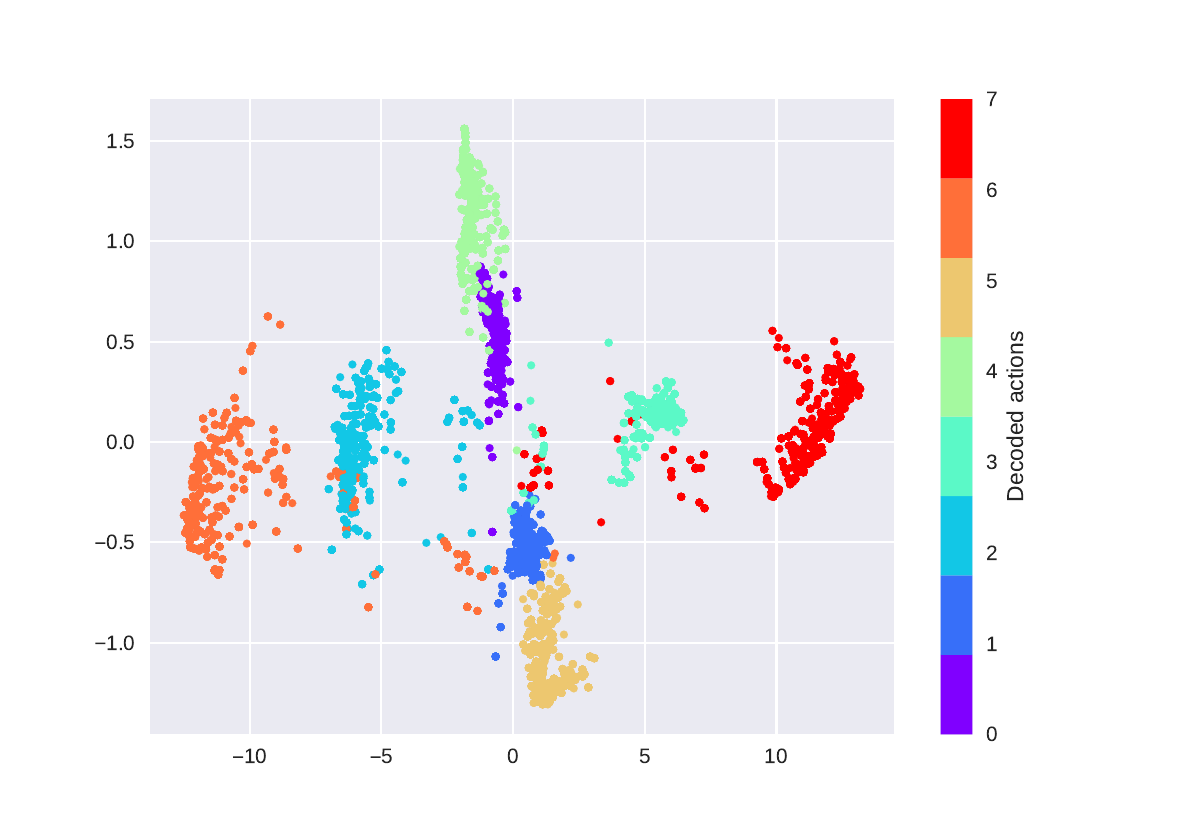}
\captionsetup{justification=centering}
    \caption{Maze $14\times14$, latent transitions}
\label{fig:maze14t}
    \end{subfigure} 
\caption{First two principal components of the learned action space $\mathcal{\bar{A}}$ and the $\Delta \bar{\text{T}}$ in the $14 \times 14$ grid-world in Figure \ref{fig:env-2obj}.}
    \label{fig:grid-envs-latent_act}
    \end{figure*}

In Figure \ref{fig:grid-envs-latent_act}, we also show the learned action representations (Figure \ref{fig:maze14a}) and learned transitions $\Delta \bar{\text{T}}$ (Figure \ref{fig:maze14t}) in the $14 \times 14$ mazes in Figure \ref{fig:grid_envs}\footnote{Again the results in the $6\times6$ grid-world are shown in Appendix B (Figure \ref{fig:grid-envs-latent_act6x6}).}. The latent transitions resemble the true transitions of the agent in the grid-world.

\subsubsection{Comparison of the Learned Policies}

\begin{figure*}[ht!]
\centering
    \begin{subfigure}{0.33\textwidth}
\includegraphics[width=1.0\linewidth]{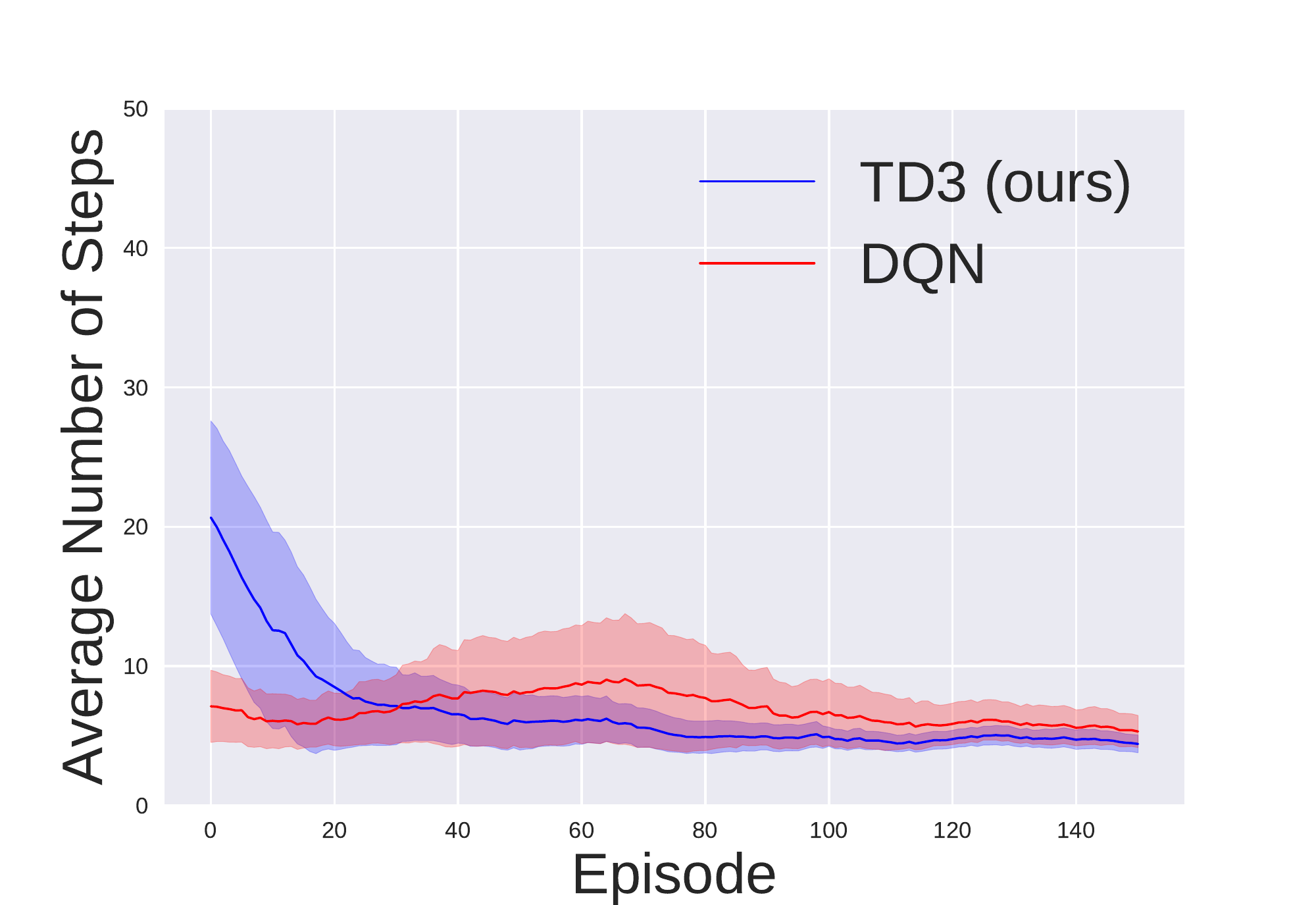}
\captionsetup{justification=centering}
    \caption{Maze $6\times6$, 4 actions}
\label{fig:pi6x6}
    \end{subfigure} \hfill
    \begin{subfigure}{0.33\textwidth}
\includegraphics[width=1.0\linewidth]{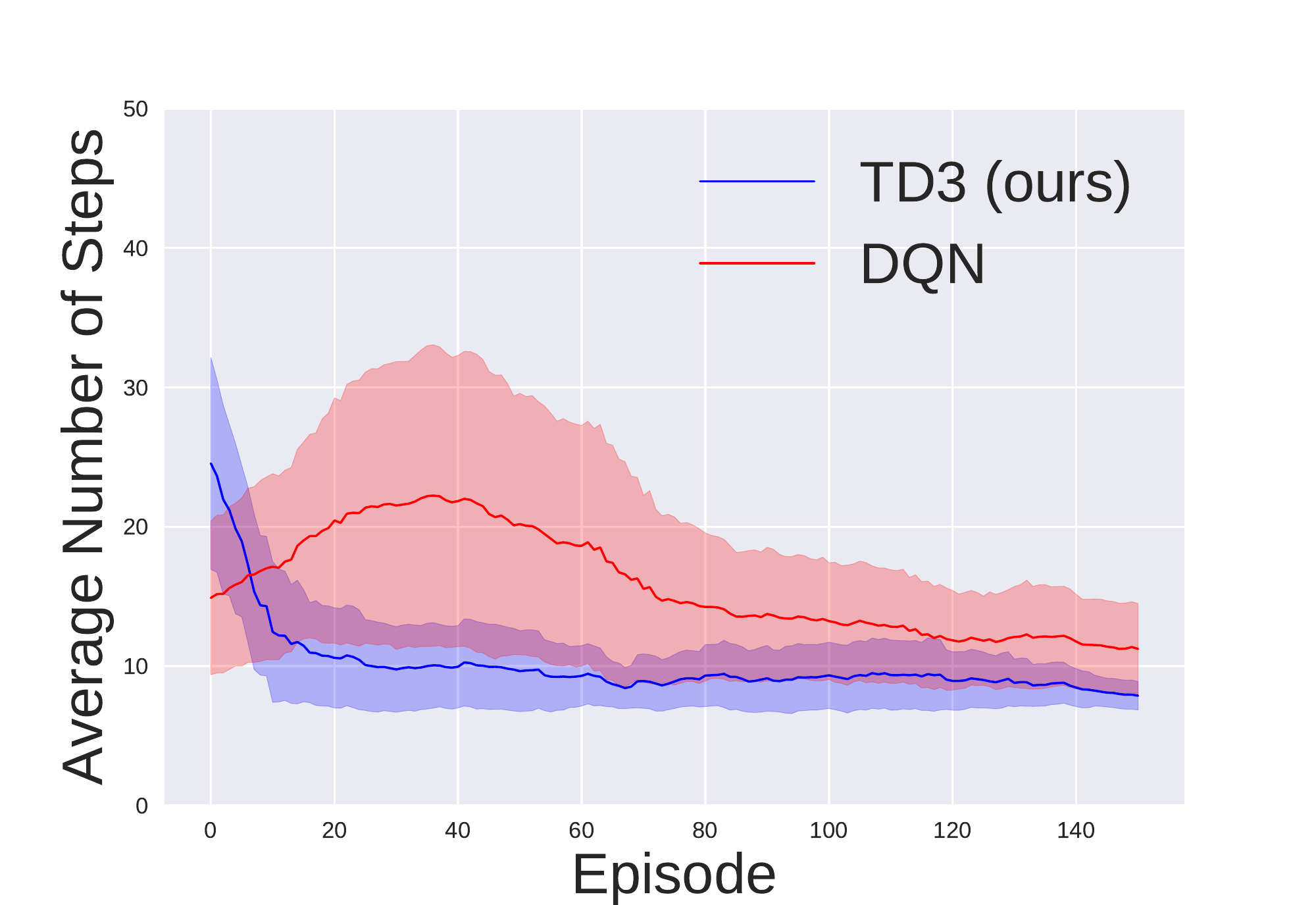}
\captionsetup{justification=centering}
    \caption{Maze $14\times14$, 8 actions}
\label{fig:pi14x14}
    \end{subfigure} \hfill
\begin{subfigure}{0.33\textwidth}
\includegraphics[width=1.0\linewidth]{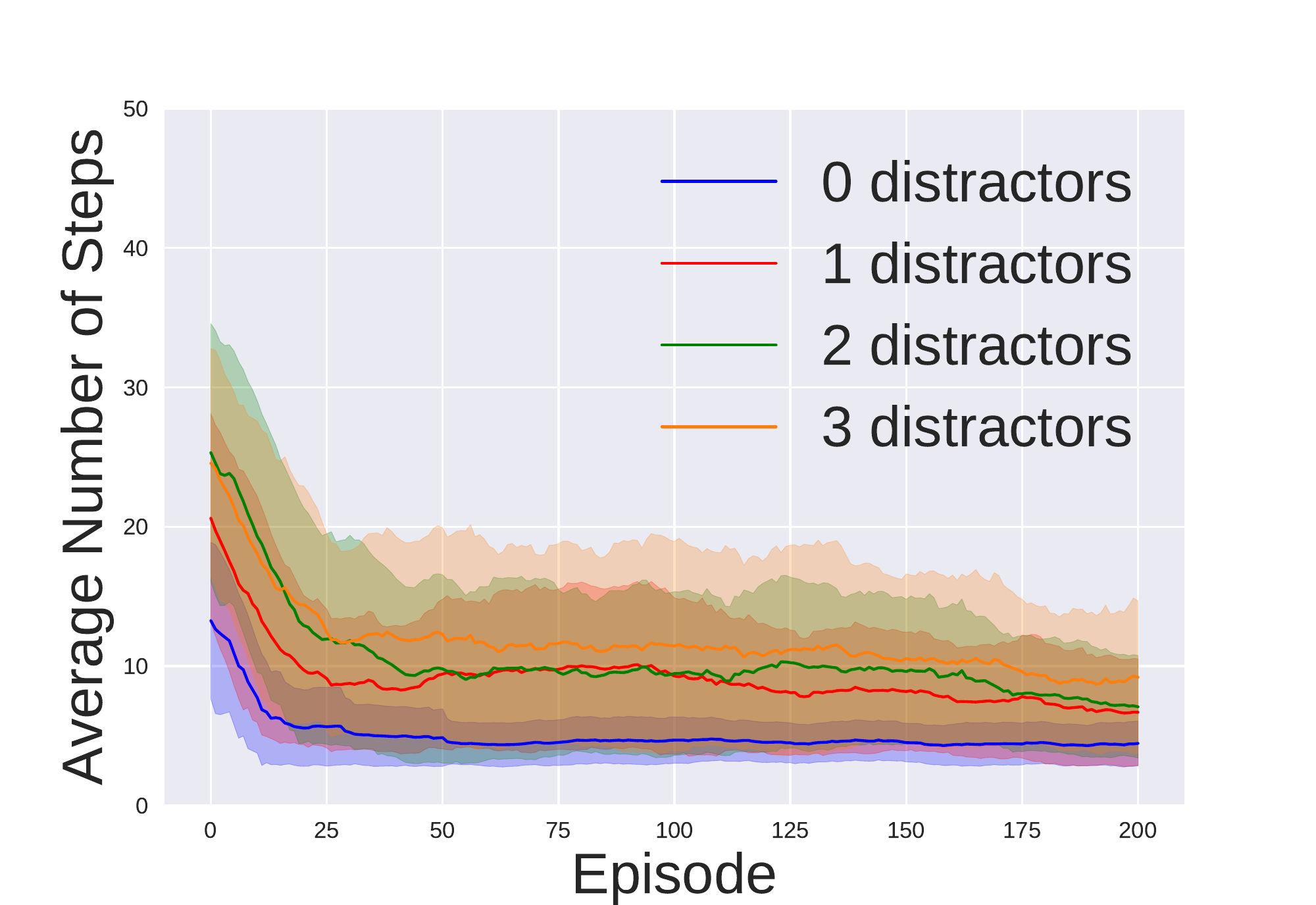}
\captionsetup{justification=centering}
    \caption{Maze $5\times5$, 4 actions, 0,1,2,3 distractors}
\label{fig:pi5x5}
    \end{subfigure} 
\caption{Average number of training steps. The solid line represents the mean and the shaded area, the variance of the best performing three random seeds out of ten.}
    \label{fig:grid_results}
    \end{figure*}

In Figure \ref{fig:grid_results}, the performance of the two agents is compared for the different grid-worlds. As soon as the state and action space grows, e.g. maze 14x14 with eight actions, the latent policy outperforms the DQN policy in terms of convergence speed to the optimal solution.

\subsection{Mobile Robot Navigation}

\subsubsection{Comparison of the  Learned Representations}

\begin{figure*}[ht!]
\centering
    \begin{subfigure}{0.33\textwidth}
\includegraphics[width=1.0\linewidth,page=4]{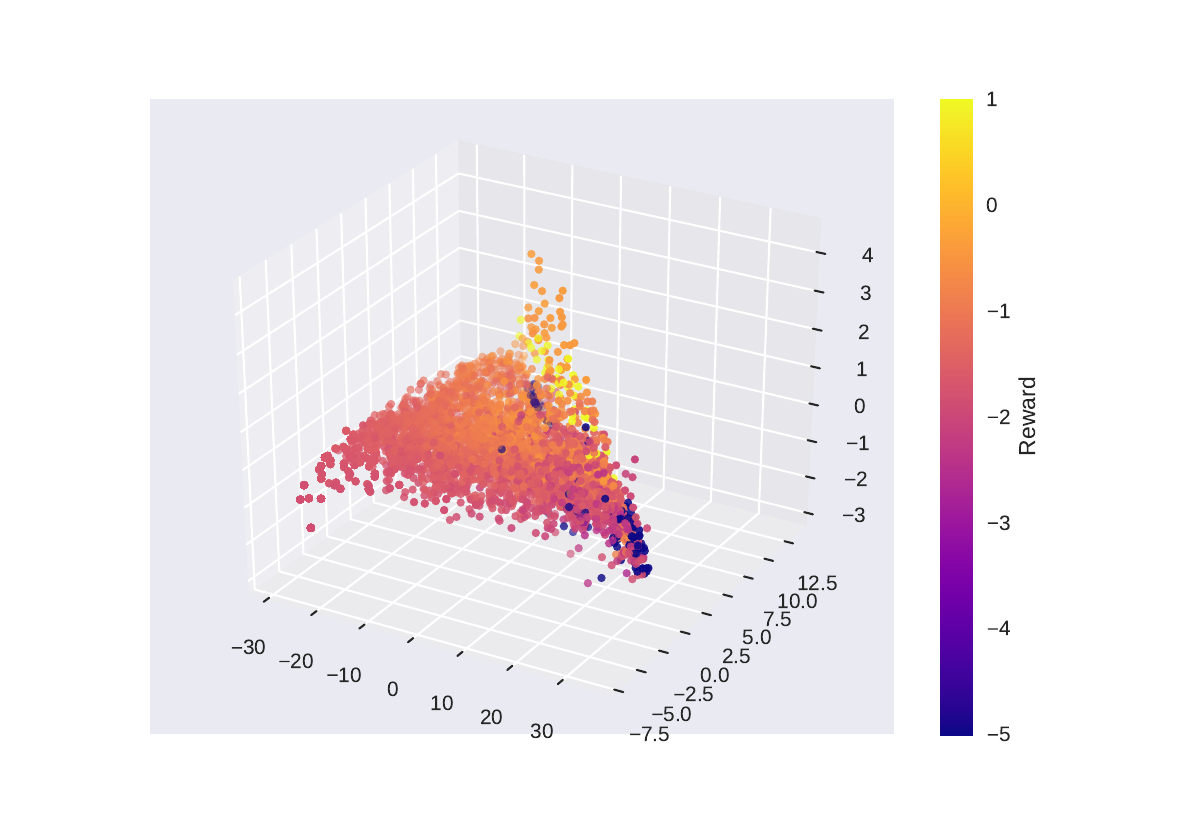}
\captionsetup{justification=centering}
    \caption{Our}
\label{fig:ourrobot}
    \end{subfigure} \hfill
    \begin{subfigure}{0.33\textwidth}
\includegraphics[width=1.0\linewidth,page=4]{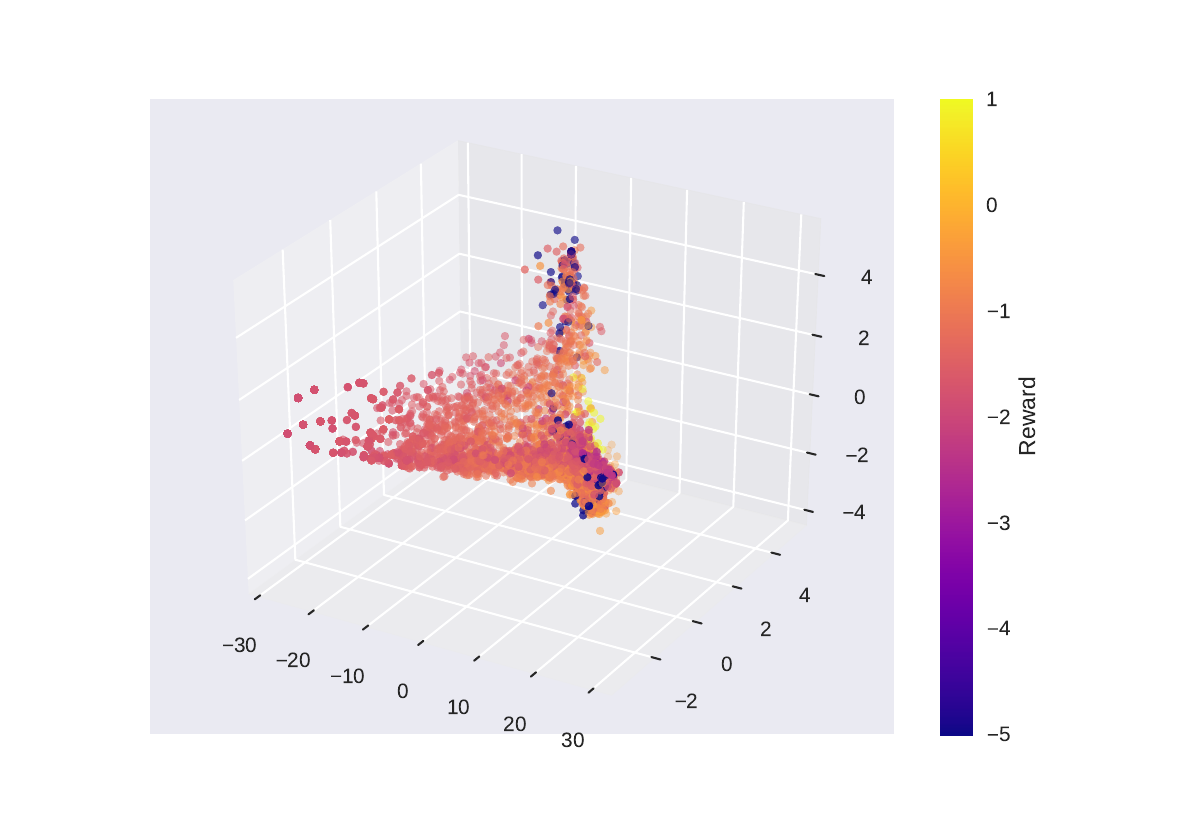}
\captionsetup{justification=centering}
    \caption{MDP-H \cite{VanderPol2020}}
\label{fig:planmdprobot}
    \end{subfigure} \hfill
\begin{subfigure}{0.33\textwidth}
\includegraphics[width=1.0\linewidth,page=4]{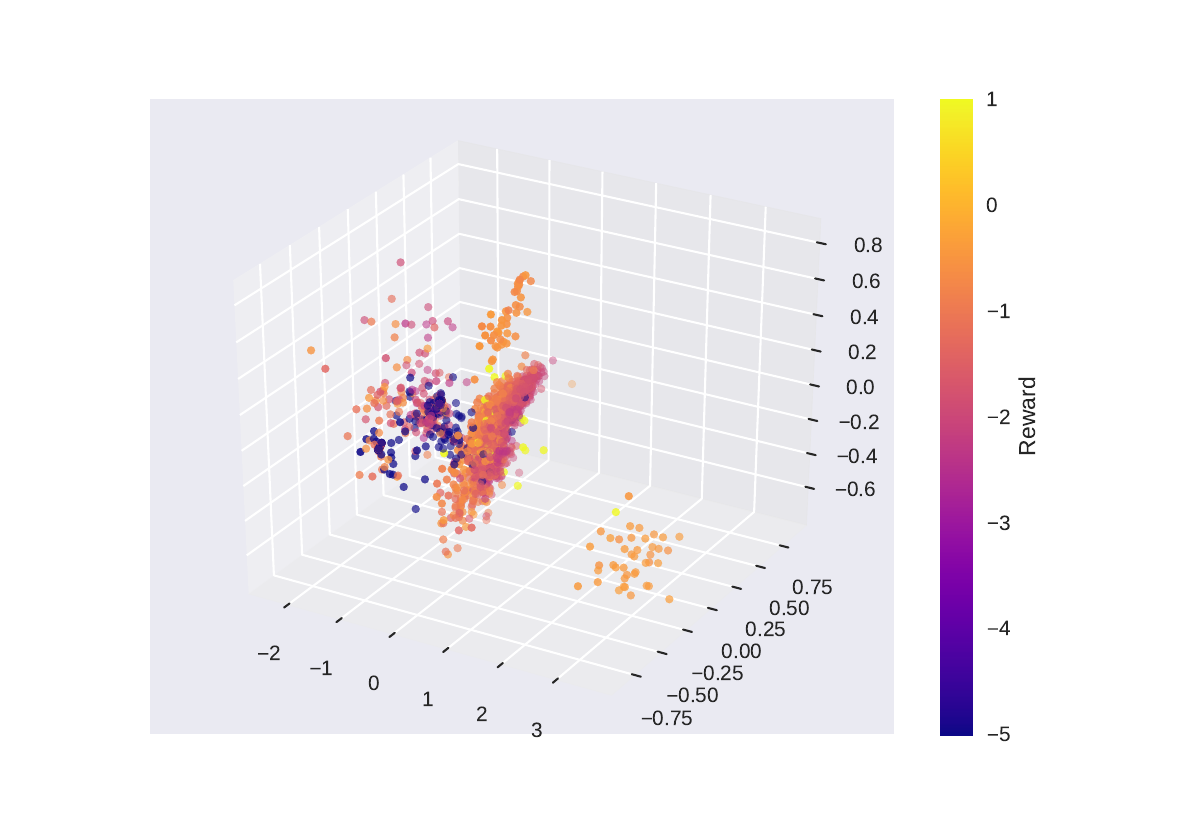}
\captionsetup{justification=centering}
    \caption{D-MDP \cite{gelada2019deepmdp}}
\label{fig:deepmdprobot}
    \end{subfigure} \hfill
    \begin{subfigure}{0.33\textwidth}
\includegraphics[width=1.0\linewidth,page=4]{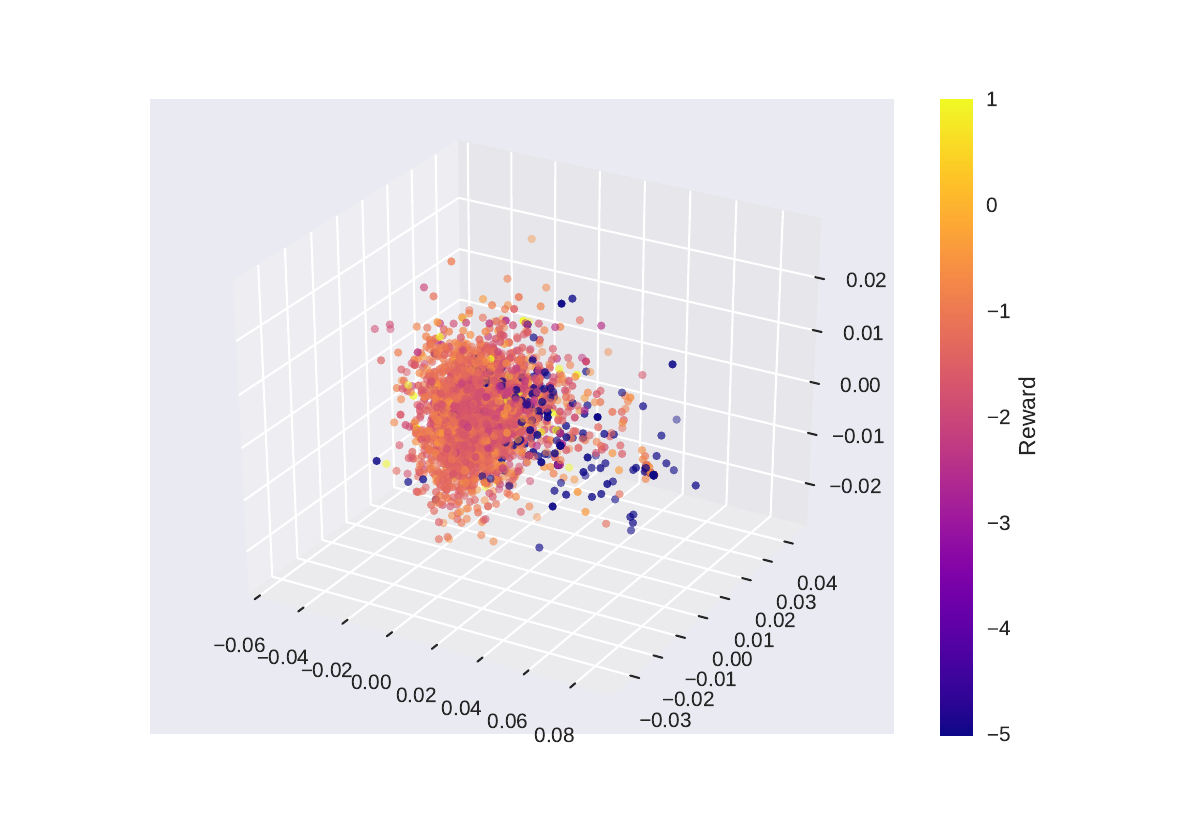}
\captionsetup{justification=centering}
    \caption{JSAE \cite{pritz2020joint}}
\label{fig:jsaerobot}
    \end{subfigure}
        \begin{subfigure}{0.33\textwidth}
\includegraphics[width=1.0\linewidth,page=4]{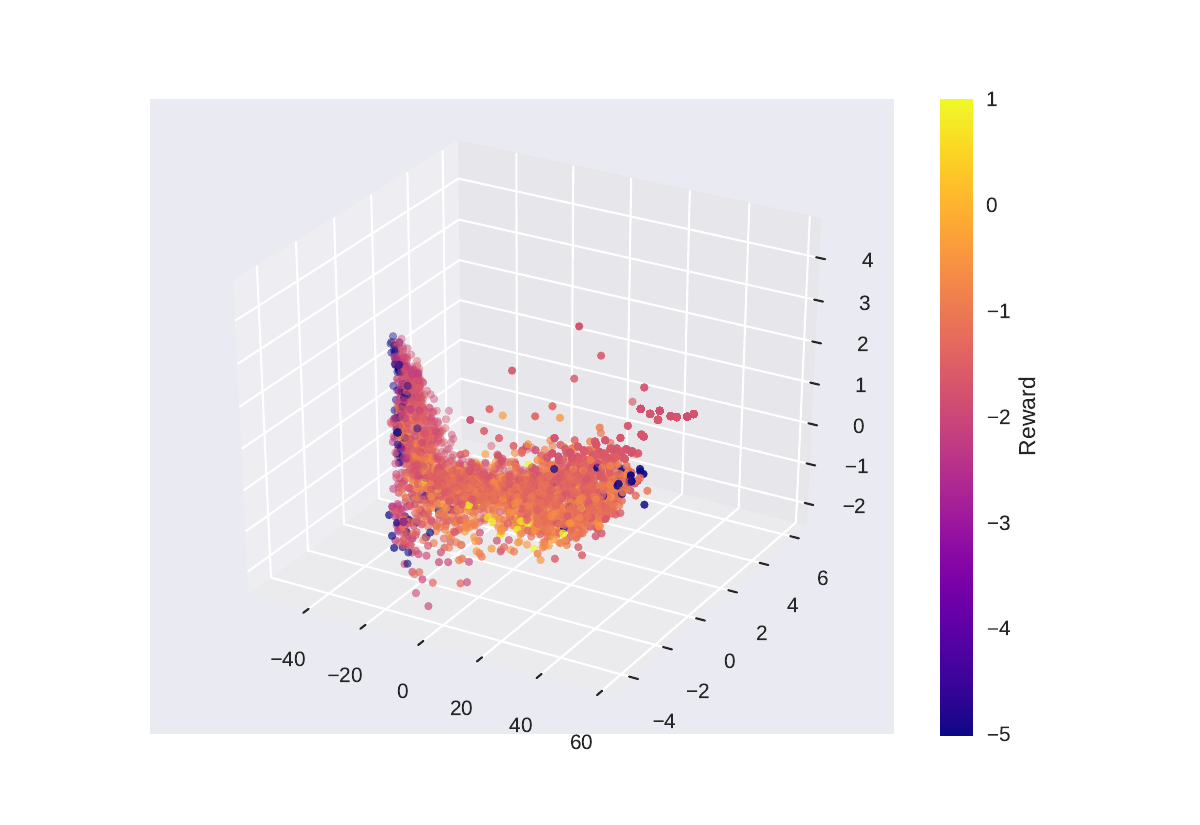}
\captionsetup{justification=centering}
    \caption{JSAE-C \cite{pritz2020joint}}
\label{fig:jsaecrobot}
    \end{subfigure}
\caption{First 3 principal components of learned state representations using the samples from the environment in Figure \ref{fig:robot_envs}}
    \label{fig:state_representation_robot}
    \end{figure*}

The learned state representations obtained with the different approaches are shown in Figure \ref{fig:state_representation_robot}. Similarly to the grid-world case, our approach can learn a valid state representation resembling the underlying state space in terms of smoothness and reward properties (i.e. distance to the target). This aspect can be noticed from the state distribution and its colour gradient in Figure \ref{fig:ourrobot}. It is worth highlighting the benefits of the contrastive loss (Equation (\ref{contrastive_loss}) for learning state representations. Especially in the case of an underlying continuous state space, such as in the robot navigation experiments, the methods employing a contrastive loss tend to improve the quality of the learned representation, and this can be noticed from Figure \ref{fig:ourrobot}, \ref{fig:planmdprobot}, and \ref{fig:jsaecrobot}.

\subsubsection{Comparison of the Learned Policies}

\begin{figure*}[ht!]
\centering
    \begin{subfigure}{0.33\textwidth}
\includegraphics[width=1.0\linewidth]{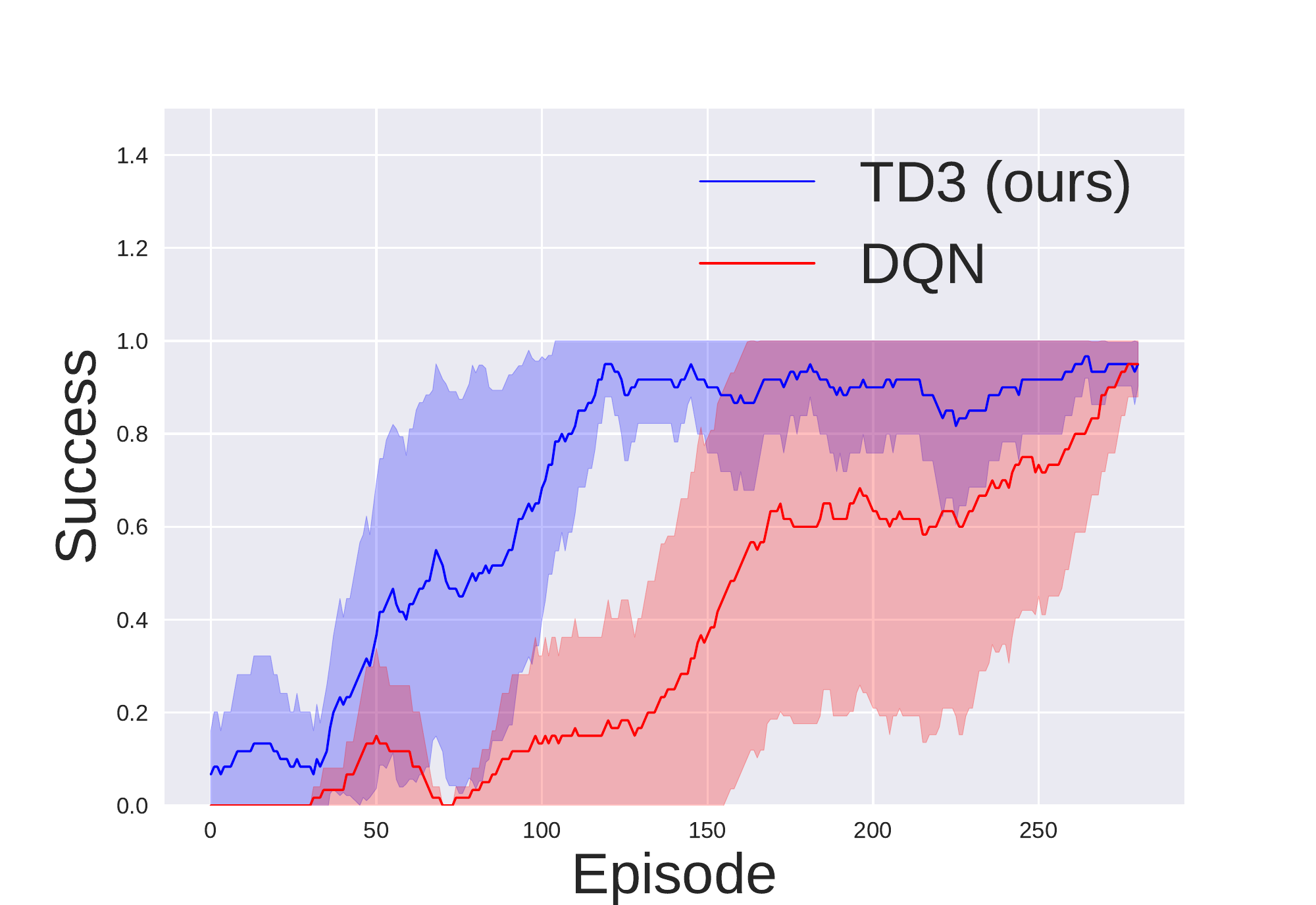}
\captionsetup{justification=centering}
    \caption{Robot, 3 actions}
\label{fig:robot3}
    \end{subfigure}
    \begin{subfigure}{0.33\textwidth}
\includegraphics[width=1.0\linewidth]{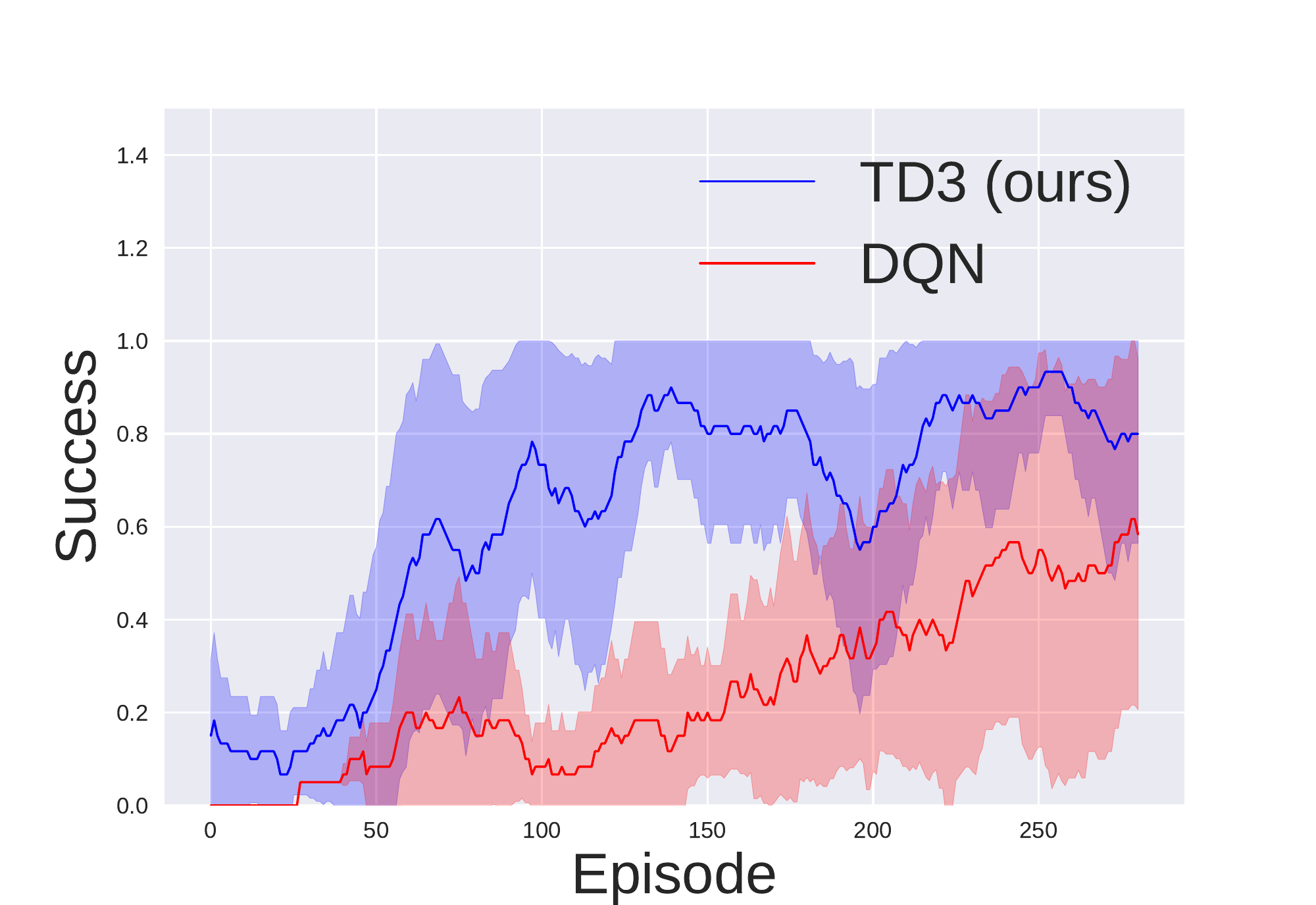}
\captionsetup{justification=centering}
    \caption{Robot, 8 actions}
\label{fig:robot8}
    \end{subfigure} 
\caption{Success ratio over training. The solid line represents the mean and the shaded area the variance of the best performing 3 random seeds out of 10.}
    \label{fig:robot_results}
    \end{figure*}

In Figure \ref{fig:robot_results}, the performance of the two agents is compared for the different action spaces. In both cases, the latent policy outperforms the DQN policy in terms of the average success ratio over training.

\section{Discussion and Future Work}\label{sec:Discussion}

We presented a framework for the self-supervised learning of state and action representations for Reinforcement Learning for high-dimensional problems. Instead of learning the complex policy $\pi_o:\mathcal{O} \longrightarrow \mathcal{A}$ mapping the observation space $\mathcal{O}$ directly to the action space $\mathcal{A}$, using self-supervised objectives (Equation (\ref{transition_loss})-(\ref{contrastive_loss}), (\ref{decoder_loss_discrete})), we transform a (potentially) high-dimensional MDP $\mathcal{M}$ (either discrete or continuous) in a homomorphic, continuous, and low-dimensional MDP $\mathcal{\bar{M}}$ in $\mathcal{\bar{S}}$ and $\mathcal{\bar{A}}$. The latent policy $\bar{\pi}: \mathcal{\bar{S}} \longrightarrow \mathcal{\bar{A}}$ is now a continuous policy, independent of the dimensions of the observation space $\mathcal{O}$ and the action space $\mathcal{A}$. Therefore, the method scales well  with the dimension of the  underlying true state, the observation space, and the action space. The latent policy can be quickly and efficiently optimised by any policy gradient algorithm.  Moreover, because the policy is learned using a state representation, it is naturally more robust against noise, disturbances, and untrained features (see Figure \ref{fig:pi5x5}). 

The proposed framework learns state and action representations through learning the MDP dynamics (transition and reward models). The framework directly combines model-free and model-based Reinforcement Learning. The learned latent transition and reward model can be used for sampling and planning. The balance between the use of the models, the real samples for exploration of the spaces and optimisation of the policy is an interesting future direction. 

The action encoder model $\psi_e$ is trained to optimise two objectives (Equation (\ref{total_loss_MDP}), (\ref{decoder_loss_discrete})), however, similarly to \cite{makhzani2015adversarial}, if priors are available, it is possible to use them to shape the learned action space and consequently the whole state-action representation.

We have only considered simple one-to-one mappings $\psi_e$ and $\delta_d$ between actions and latent actions. However, it is possible to learn high-level action representations by learning many-to-one mappings. A sequence of actions is mapped to a single latent action, and the latent action is consequently decoded into a sequence of actions back. In this case, it is possible to rely on the semi-MDP \cite{sutton1999between} framework to learn a semi-MDP homomorphism. This aspect may have an impact in all the robotics applications with complex action spaces or in natural language processing. 

Eventually, we have only considered the case of deterministic MDPs, but the framework can be extended to stochastic MDPs. Moreover, we have restricted the study to Markovian observation  space so that we could rely on the MDP framework. However an important future step is to bring this framework to partially-observable MDPs (POMDPs), in which a single observation is not sufficient to unequivocally determine the agent's state. In this context, it is interesting to investigate the use of recurrent architectures and transformers \cite{vaswani2017attention}. 

\section{Conclusion}\label{sec:Conclusion}

In this paper, we proposed a framework for state and action representation learning for Reinforcement Learning. Our approach transforms a given MDP $\mathcal{M}$ into an homomorphic MDP $\mathcal{\bar{M}}$. The new MDP $\mathcal{\bar{M}}=\langle \mathcal{\bar{S}}, \mathcal{\bar{A}}, \bar{\text{T}}, \bar{\text{R}}\rangle$ has continuous state and action spaces, but it is easier to solve using any policy gradient algorithms. We showed that the optimal latent policy $\bar{\pi}$ for $\mathcal{\bar{M}}$ is optimal for the original MDP $\mathcal{M}$ and that it can be efficiently and effectively learned. The optimal latent policy converges faster than the DQN agent trained on a state representation to the optimal solution as soon as the underlying true state and action spaces grow in size and complexity.  

\bibliographystyle{unsrt}  
\bibliography{references} 

\begin{thebibliography}{10}

\bibitem{sutton_reinforcement_2018}
Richard~S. Sutton and Andrew~G. Barto.
\newblock {\em Reinforcement learning: an introduction}.
\newblock Adaptive computation and machine learning series. The MIT Press,
  Cambridge, Massachusetts, second edition edition, 2018.

\bibitem{mnih2013playing}
Volodymyr Mnih, Koray Kavukcuoglu, David Silver, Alex Graves, Ioannis
  Antonoglou, Daan Wierstra, and Martin Riedmiller.
\newblock Playing atari with deep reinforcement learning.
\newblock {\em arXiv preprint arXiv:1312.5602}, 2013.

\bibitem{kober2013reinforcement}
Jens Kober, J~Andrew Bagnell, and Jan Peters.
\newblock Reinforcement learning in robotics: A survey.
\newblock {\em The International Journal of Robotics Research},
  32(11):1238--1274, 2013.

\bibitem{SRL_survey}
Timothee Lesort, Natalia Diaz-Rodríguez, Jean-Franois Goudou, and David
  Filliat.
\newblock State representation learning for control: An overview.
\newblock {\em Neural Networks}, 2018.

\bibitem{gelada2019deepmdp}
Carles Gelada, Saurabh Kumar, Jacob Buckman, Ofir Nachum, and Marc~G.
  Bellemare.
\newblock Deepmdp: Learning continuous latent space models for representation
  learning, 2019.

\bibitem{francoislavet2018combined}
Vincent François-Lavet, Yoshua Bengio, Doina Precup, and Joelle Pineau.
\newblock Combined reinforcement learning via abstract representations, 2018.

\bibitem{VanderPol2020}
Elise van~der Pol, Thomas Kipf, Frans~A. Oliehoek, and Max Welling.
\newblock {Plannable Approximations to MDP Homomorphisms: Equivariance under
  Actions}.
\newblock (Aamas), 2020.

\bibitem{dulacarnold2015deep}
Gabriel Dulac-Arnold, Richard Evans, Hado van Hasselt, Peter Sunehag, Timothy
  Lillicrap, Jonathan Hunt, Timothy Mann, Theophane Weber, Thomas Degris, and
  Ben Coppin.
\newblock Deep reinforcement learning in large discrete action spaces, 2015.

\bibitem{losey2019controlling}
Dylan~P. Losey, Krishnan Srinivasan, Ajay Mandlekar, Animesh Garg, and Dorsa
  Sadigh.
\newblock Controlling assistive robots with learned latent actions, 2019.

\bibitem{ch2019learning}
Yash Chandak, Georgios Theocharous, James Kostas, Scott Jordan, and Philip~S.
  Thomas.
\newblock Learning action representations for reinforcement learning, 2019.

\bibitem{ravindran2001symmetries}
Balaraman Ravindran and Andrew~G Barto.
\newblock Symmetries and model minimization in markov decision processes, 2001.

\bibitem{ravindran2004approximate}
Balaraman Ravindran and Andrew~G Barto.
\newblock Approximate homomorphisms: A framework for non-exact minimization in
  markov decision processes.
\newblock 2004.

\bibitem{fujimoto2018addressing}
Scott Fujimoto, Herke Hoof, and David Meger.
\newblock Addressing function approximation error in actor-critic methods.
\newblock In {\em International Conference on Machine Learning}, pages
  1587--1596. PMLR, 2018.

\bibitem{van2016deep}
Hado Van~Hasselt, Arthur Guez, and David Silver.
\newblock Deep reinforcement learning with double q-learning.
\newblock In {\em Proceedings of the AAAI Conference on Artificial
  Intelligence}, volume~30, 2016.

\bibitem{lillicrap2015continuous}
Timothy~P Lillicrap, Jonathan~J Hunt, Alexander Pritzel, Nicolas Heess, Tom
  Erez, Yuval Tassa, David Silver, and Daan Wierstra.
\newblock Continuous control with deep reinforcement learning.
\newblock {\em arXiv preprint arXiv:1509.02971}, 2015.

\bibitem{taylor2008bounding}
Jonathan Taylor, Doina Precup, and Prakash Panagaden.
\newblock Bounding performance loss in approximate mdp homomorphisms.
\newblock {\em Advances in Neural Information Processing Systems},
  21:1649--1656, 2008.

\bibitem{finn2016deep}
Chelsea Finn, Xin~Yu Tan, Yan Duan, Trevor Darrell, Sergey Levine, and Pieter
  Abbeel.
\newblock Deep spatial autoencoders for visuomotor learning, 2016.

\bibitem{mattner2012learn}
Jan Mattner, Sascha Lange, and Martin Riedmiller.
\newblock Learn to swing up and balance a real pole based on raw visual input
  data.
\newblock In {\em International Conference on Neural Information Processing},
  pages 126--133. Springer, 2012.

\bibitem{van2016stable}
Herke Van~Hoof, Nutan Chen, Maximilian Karl, Patrick van~der Smagt, and Jan
  Peters.
\newblock Stable reinforcement learning with autoencoders for tactile and
  visual data.
\newblock In {\em 2016 IEEE/RSJ international conference on intelligent robots
  and systems (IROS)}, pages 3928--3934. IEEE, 2016.

\bibitem{de2018integrating}
Tim de~Bruin, Jens Kober, Karl Tuyls, and Robert Babu{\v{s}}ka.
\newblock Integrating state representation learning into deep reinforcement
  learning.
\newblock {\em IEEE Robotics and Automation Letters}, 3(3):1394--1401, 2018.

\bibitem{wahlstrom2015pixels}
Niklas Wahlstr{\"o}m, Thomas~B Sch{\"o}n, and Marc~Peter Deisenroth.
\newblock From pixels to torques: Policy learning with deep dynamical models.
\newblock {\em arXiv preprint arXiv:1502.02251}, 2015.

\bibitem{Francois-Lavet2019}
Vincent Francois-Lavet, Yoshua Bengio, Doina Precup, and Joelle Pineau.
\newblock {Combined Reinforcement Learning via Abstract Representations}.
\newblock {\em Proceedings of the AAAI Conference on Artificial Intelligence},
  33:3582--3589, 2019.

\bibitem{jonschkowski2015learning}
Rico Jonschkowski and Oliver Brock.
\newblock Learning state representations with robotic priors.
\newblock {\em Autonomous Robots}, 39(3):407--428, 2015.

\bibitem{jonschkowski2017pves}
Rico Jonschkowski, Roland Hafner, Jonathan Scholz, and Martin Riedmiller.
\newblock Pves: Position-velocity encoders for unsupervised learning of
  structured state representations.
\newblock {\em arXiv preprint arXiv:1705.09805}, 2017.

\bibitem{botteghi2020low}
Nicolò Botteghi, Ruben Obbink, Daan Geijs, Mannes Poel, Beril Sirmacek,
  Christoph Brune, Abeje Mersha, and Stefano Stramigioli.
\newblock Low dimensional state representation learning with reward-shaped
  priors, 2020.

\bibitem{sutton1999between}
Richard~S Sutton, Doina Precup, and Satinder Singh.
\newblock Between mdps and semi-mdps: A framework for temporal abstraction in
  reinforcement learning.
\newblock {\em Artificial intelligence}, 112(1-2):181--211, 1999.

\bibitem{stolle2002learning}
Martin Stolle and Doina Precup.
\newblock Learning options in reinforcement learning.
\newblock In {\em International Symposium on abstraction, reformulation, and
  approximation}, pages 212--223. Springer, 2002.

\bibitem{dietterich1998maxq}
Thomas~G Dietterich.
\newblock The maxq method for hierarchical reinforcement learning.
\newblock In {\em ICML}, volume~98, pages 118--126. Citeseer, 1998.

\bibitem{vezhnevets2017feudal}
Alexander~Sasha Vezhnevets, Simon Osindero, Tom Schaul, Nicolas Heess, Max
  Jaderberg, David Silver, and Koray Kavukcuoglu.
\newblock Feudal networks for hierarchical reinforcement learning.
\newblock {\em arXiv preprint arXiv:1703.01161}, 2017.

\bibitem{pritz2020joint}
Paul~J Pritz, Liang Ma, and Kin~K Leung.
\newblock Joint state-action embedding for efficient reinforcement learning.
\newblock {\em arXiv preprint arXiv:2010.04444}, 2020.

\bibitem{kipf2019contrastive}
Thomas Kipf, Elise van~der Pol, and Max Welling.
\newblock Contrastive learning of structured world models.
\newblock {\em arXiv preprint arXiv:1911.12247}, 2019.

\bibitem{gym_minigrid}
Maxime Chevalier-Boisvert, Lucas Willems, and Suman Pal.
\newblock Minimalistic gridworld environment for openai gym.
\newblock \url{https://github.com/maximecb/gym-minigrid}, 2018.

\bibitem{rohmer2013v}
Eric Rohmer, Surya~PN Singh, and Marc Freese.
\newblock V-rep: A versatile and scalable robot simulation framework.
\newblock In {\em 2013 IEEE/RSJ International Conference on Intelligent Robots
  and Systems}, pages 1321--1326. IEEE, 2013.

\bibitem{james2019pyrep}
Stephen James, Marc Freese, and Andrew~J Davison.
\newblock Pyrep: Bringing v-rep to deep robot learning.
\newblock {\em arXiv preprint arXiv:1906.11176}, 2019.

\bibitem{mnih2015human}
Volodymyr Mnih, Koray Kavukcuoglu, David Silver, Andrei~A Rusu, Joel Veness,
  Marc~G Bellemare, Alex Graves, Martin Riedmiller, Andreas~K Fidjeland, Georg
  Ostrovski, et~al.
\newblock Human-level control through deep reinforcement learning.
\newblock {\em nature}, 518(7540):529--533, 2015.

\bibitem{jaderberg2016reinforcement}
Max Jaderberg, Volodymyr Mnih, Wojciech~Marian Czarnecki, Tom Schaul, Joel~Z
  Leibo, David Silver, and Koray Kavukcuoglu.
\newblock Reinforcement learning with unsupervised auxiliary tasks.
\newblock {\em arXiv preprint arXiv:1611.05397}, 2016.

\bibitem{makhzani2015adversarial}
Alireza Makhzani, Jonathon Shlens, Navdeep Jaitly, Ian Goodfellow, and Brendan
  Frey.
\newblock Adversarial autoencoders.
\newblock {\em arXiv preprint arXiv:1511.05644}, 2015.

\bibitem{vaswani2017attention}
Ashish Vaswani, Noam Shazeer, Niki Parmar, Jakob Uszkoreit, Llion Jones,
  Aidan~N Gomez, Lukasz Kaiser, and Illia Polosukhin.
\newblock Attention is all you need.
\newblock {\em arXiv preprint arXiv:1706.03762}, 2017.

\end{thebibliography}

\section*{Appendix A}\label{appendixA}

\textbf{\textit{Proposition 2:}} For all deterministic functions $\delta_d$, the gradient  $\nabla_{\pmb{\theta}_{\bar{\pi}}} J_{\bar{\pi}}(\pmb{\theta}_{\bar{\pi}})$ of the performance measure of the latent policy $\bar{\pi}:\mathcal{\bar{S}} \longrightarrow \mathcal{\bar{A}}$ is equivalent to the gradient $\nabla_{\pmb{\theta}_{\bar{\pi}}} J_{\pi_i}(\pmb{\theta}_{\bar{\pi}}, \pmb{\theta}_{\delta_d})$ of the performance measure the intermediate policy  $\pi_i:\mathcal{\bar{S}} \longrightarrow \mathcal{A}$:

\begin{equation}
    \nabla_{\pmb{\theta}_{\bar{\pi}}} J_{\pi_i}(\pmb{\theta}_{\bar{\pi}}, \pmb{\theta}_{\delta_d}) = \nabla_{\pmb{\theta}_{\bar{\pi}}} J_{\bar{\pi}}(\pmb{\theta}_{\bar{\pi}}) 
\end{equation}

\textbf{\textit{Proof}} (adapted from \cite{ch2019learning})

Given the relation between the intermediate policy $\pi_i$ and the latent policy $\bar{\pi}$:

\begin{equation}
    \pi_{i}(a | \bar{s}):=\int_{\delta_d^{-1}(a)} \bar{\pi}(\bar{a} | \bar{s}) \mathrm{d} \bar{a}
\end{equation}

We can express the performance measure of the intermediate policy with as:

\begin{equation}
    \begin{split}
J_{\pi_i}(\pmb{\theta}_{\bar{\pi}}, \pmb{\theta}_{\delta_d}) &=\sum_{\bar{s} \in \mathcal{\bar{S}}} d_{0}(\bar{s}) \text{V}^{\pi_{i}}(\bar{s}) \\
&=\sum_{\bar{s} \in \mathcal{\bar{S}}} d_{0}(\bar{s}) \sum_{a \in \mathcal{A}} \int_{\delta_d^{-1}(a)} \bar{\pi}(\bar{a} \mid \bar{s}) \text{Q}^{\pi_{i}}(\bar{s}, a) \mathrm{d} \bar{a}
\end{split}
\end{equation}

If we now take the gradient of the  performance measure of the intermediate policy, we obtain:

\begin{equation}
   \nabla_{\pmb{\theta}_{\bar{\pi}}} J_{\pi_i}(\pmb{\theta}_{\bar{\pi}}, \pmb{\theta}_{\delta_d}) =\nabla_{\pmb{\theta}_{\bar{\pi}}}\left[\sum_{\bar{s} \in \mathcal{\bar{S}}} d_{0}(\bar{s}) \sum_{a \in \mathcal{A}} \int_{\delta_d^{-1}(a)} \bar{\pi}(\bar{a} \mid \bar{s}) \text{Q}^{\pi_{i}}(\bar{s}, a) \mathrm{d} e\right]
   \label{gradient_intermediate_policy}
\end{equation}

Using the policy gradient theorem \cite{sutton_reinforcement_2018} for the intermediate policy $\pi_i$, we can rewrite Equation (\ref{gradient_intermediate_policy}) as:

\begin{equation}
    \begin{split}
\nabla_{\pmb{\theta}_{\bar{\pi}}} J_{\pi_i}(\pmb{\theta}_{\bar{\pi}}, \pmb{\theta}_{\delta_d})  &=\sum_{t=0}^{\infty} \mathbb{E}\left[\sum_{a \in \mathcal{A}} \gamma^{t} \text{Q}^{\pi_{i}}(\bar{S}_t, a) \nabla_{\pmb{\theta}_{\bar{\pi}}}\left(\int_{\delta_d^{-1}(a)} \bar{\pi}\left(\bar{a} \mid \bar{S}_t\right) \mathrm{d} \bar{a}\right)\right] \\
&=\sum_{t=0}^{\infty} \mathbb{E}\left[\sum_{a \in \mathcal{A}} \gamma^{t} \int_{\delta_d^{-1}(a)} \nabla_{\pmb{\theta}_{\bar{\pi}}}\left(\bar{\pi}\left(\bar{a} \mid \bar{S}_t\right)\right) \text{Q}^{\pi_{i}}(\bar{S}_t, a)  \mathrm{d} \bar{a}\right] \\
&=\sum_{t=0}^{\infty} \mathbb{E}\left[\sum_{a \in \mathcal{A}} \gamma^{t} \int_{\delta_d^{-1}(a)} \bar{\pi}\left(\bar{a} \mid \bar{S}_t\right)\nabla_{\pmb{\theta}_{\bar{\pi}}} \ln \left(\bar{\pi}\left(\bar{a} \mid \bar{S}_t\right)\right) \text{Q}^{\pi_i}(\bar{S}_t, a)  \mathrm{d} \bar{a}\right]
\end{split}
\end{equation}

Because latent actions are deterministically mapped to actions, $\text{Q}^{\pi_i}(\bar{S}_t,a)=\text{Q}^{\bar{\pi}}(\bar{S}_t,\bar{a})$. Thus:

\begin{equation}
    \nabla_{\pmb{\theta}_{\bar{\pi}}} J_{\pi_i}(\pmb{\theta}_{\bar{\pi}}, \pmb{\theta}_{\delta_d}) =\sum_{t=0}^{\infty} \mathbb{E}\left[\gamma^{t} \sum_{a \in \mathcal{A}} \int_{\delta_d^{-1}(a)} \bar{\pi}\left(\bar{a} \mid \bar{S}_t\right) \nabla_{\pmb{\theta}_{\bar{\pi}}} \ln \left(\bar{\pi}\left(\bar{a} \mid \bar{S}_{t}\right)\right) \text{Q}^{\pi_{i}}\left(\bar{S}_{t}, \bar{a}\right) \mathrm{d} \bar{a}\right]
\end{equation}

Eventually, the summation over $a$ and the integral over $\delta_d(a)$ can by replace by the integral over the domain of the latent action space $\bar{a}$. Therefore:

\begin{equation}
    \begin{split}
\nabla_{\pmb{\theta}_{\bar{\pi}}} J_{\pi_i}(\pmb{\theta}_{\bar{\pi}}, \pmb{\theta}_{\delta_d})  &=\sum_{t=0}^{\infty} \mathbb{E}\left[\gamma^{t} \int_{\bar{a}} \bar{\pi}\left(\bar{a} \mid \bar{S}_{t}\right) \nabla_{\pmb{\theta}_{\bar{\pi}}} \ln \left(\bar{\pi}\left(\bar{a} \mid \bar{S}_{t}\right)\right) \text{Q}^{\bar{\pi}}\left(\bar{S}_{t}, \bar{a}\right) \mathrm{d} \bar{a}\right] \\
&=\sum_{t=0}^{\infty} \mathbb{E}\left[\gamma^{t} \int_{\bar{a}} \text{Q}^{\bar{\pi}}\left(\bar{S}_{t}, \bar{a}\right) \nabla_{\pmb{\theta}_{\bar{\pi}}} \bar{\pi}\left(\bar{a} \mid \bar{S}_{t}\right) \mathrm{d} \bar{a}\right] \\
&=\nabla_{\pmb{\theta}_{\bar{\pi}}} J_{\bar{\pi}}(\pmb{\theta}_{\bar{\pi}})
\end{split}
\end{equation}

\section*{Appendix B}\label{appendixB}

\begin{figure*}[ht!]
\centering
    \begin{subfigure}{0.33\textwidth}
    \centering
\includegraphics[width=1.0\linewidth,page=1]{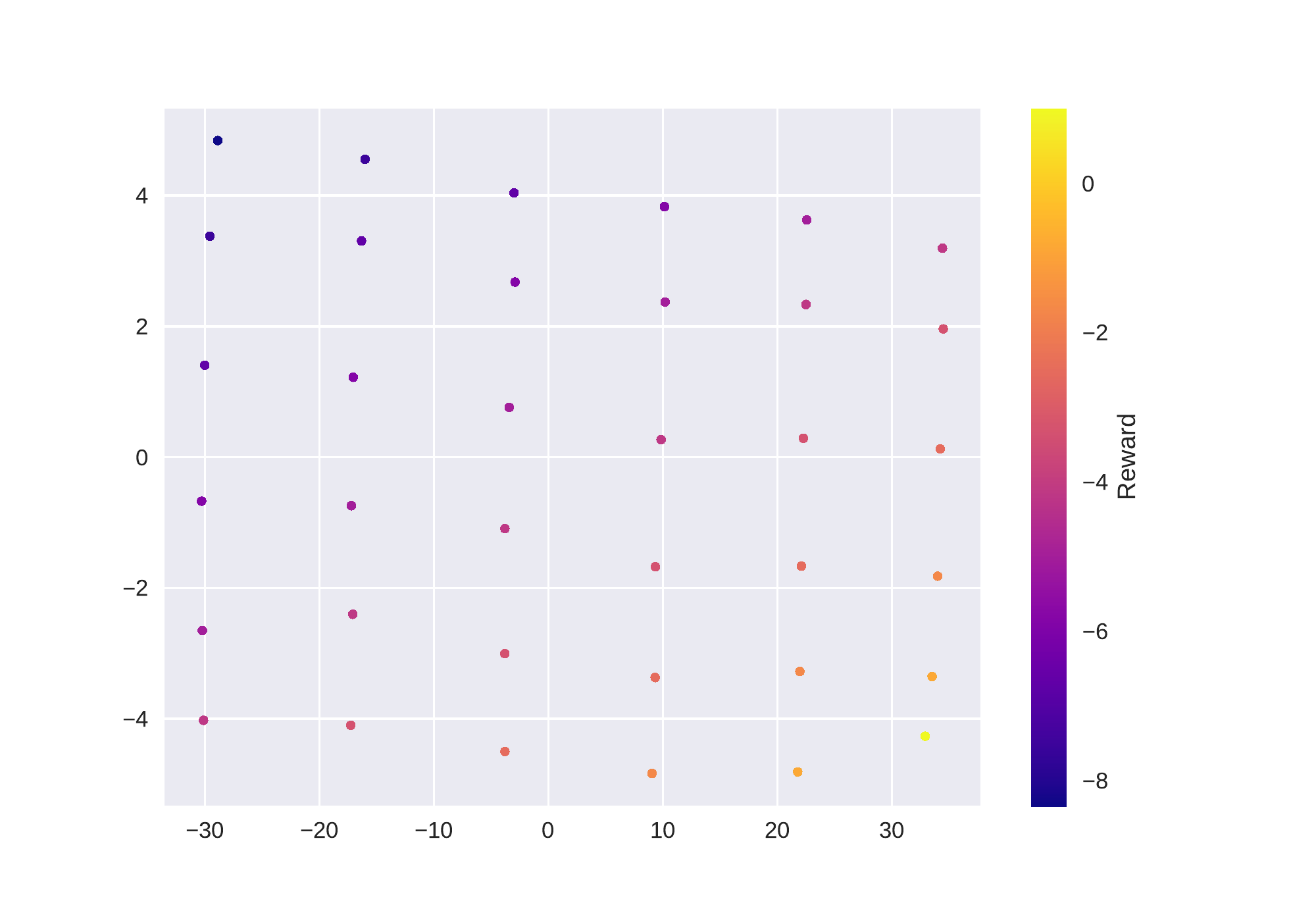}
\captionsetup{justification=centering}
    \caption{Our}
\label{fig:our}
    \end{subfigure} \hfill
    \begin{subfigure}{0.33\textwidth}
    \centering
\includegraphics[width=1.0\linewidth,page=1]{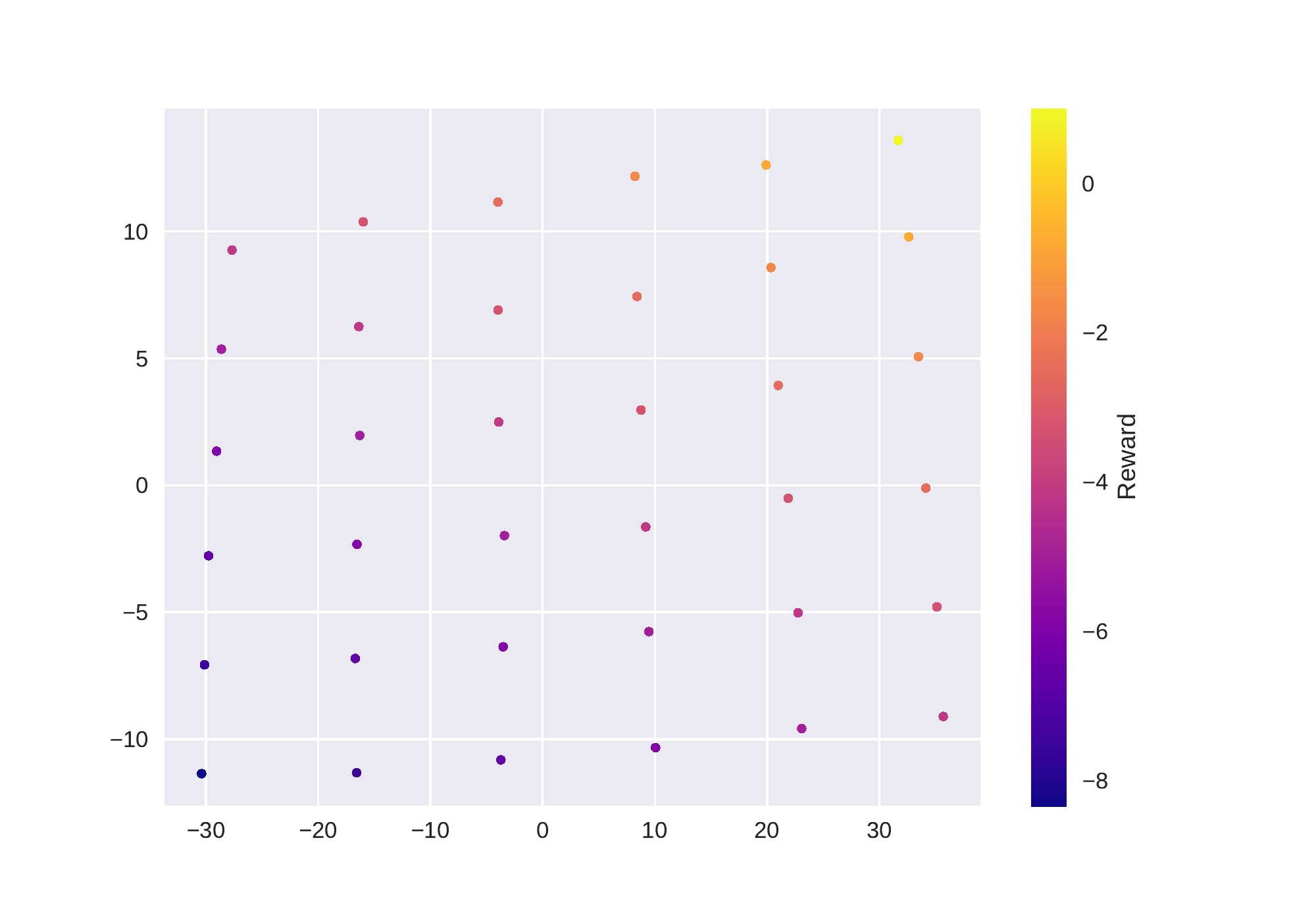}
\captionsetup{justification=centering}
    \caption{MDP-H \cite{VanderPol2020}}
\label{fig:planmdp}
    \end{subfigure} \hfill
\begin{subfigure}{0.33\textwidth}
\centering
\includegraphics[width=1.0\linewidth,page=1]{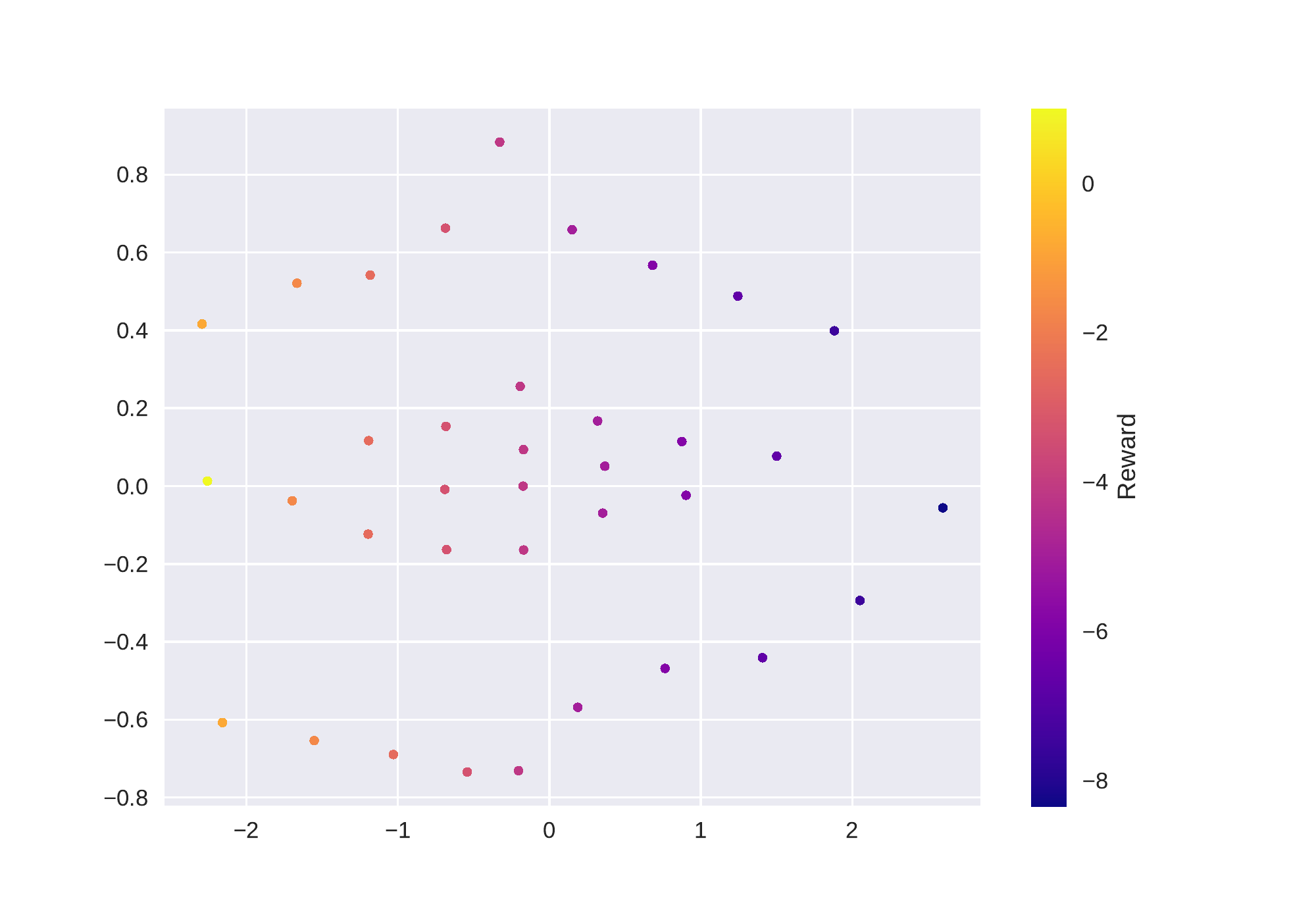}
\captionsetup{justification=centering}
    \caption{D-MDP \cite{gelada2019deepmdp}}
\label{fig:deepmdp}
    \end{subfigure} \hfill
    \begin{subfigure}{0.33\textwidth}
    \centering
\includegraphics[width=1.0\linewidth,page=1]{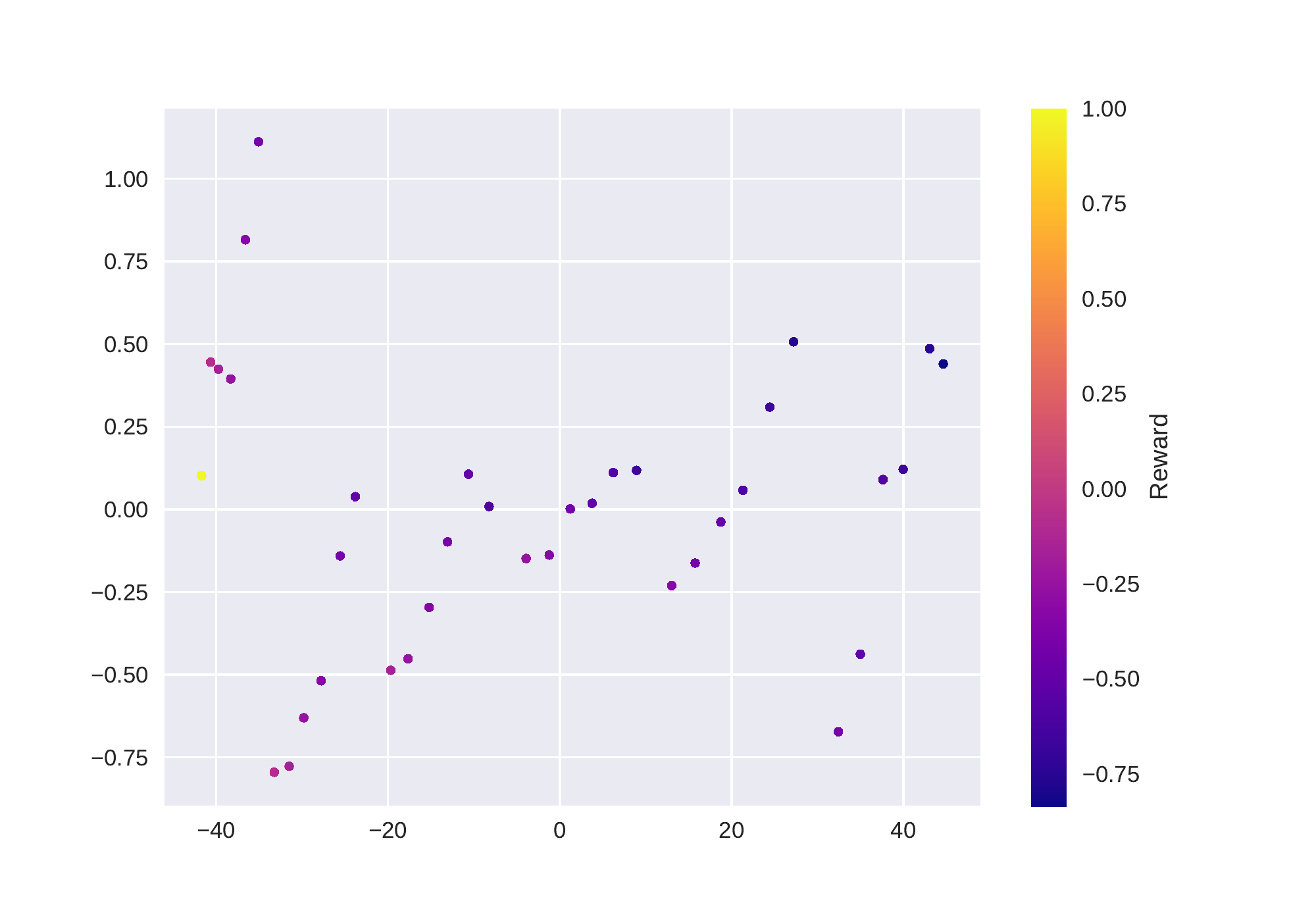}
\captionsetup{justification=centering}
    \caption{JSAE \cite{pritz2020joint}}
\label{fig:jsae}
    \end{subfigure}
        \begin{subfigure}{0.33\textwidth}
        \centering
\includegraphics[width=1.0\linewidth,page=1]{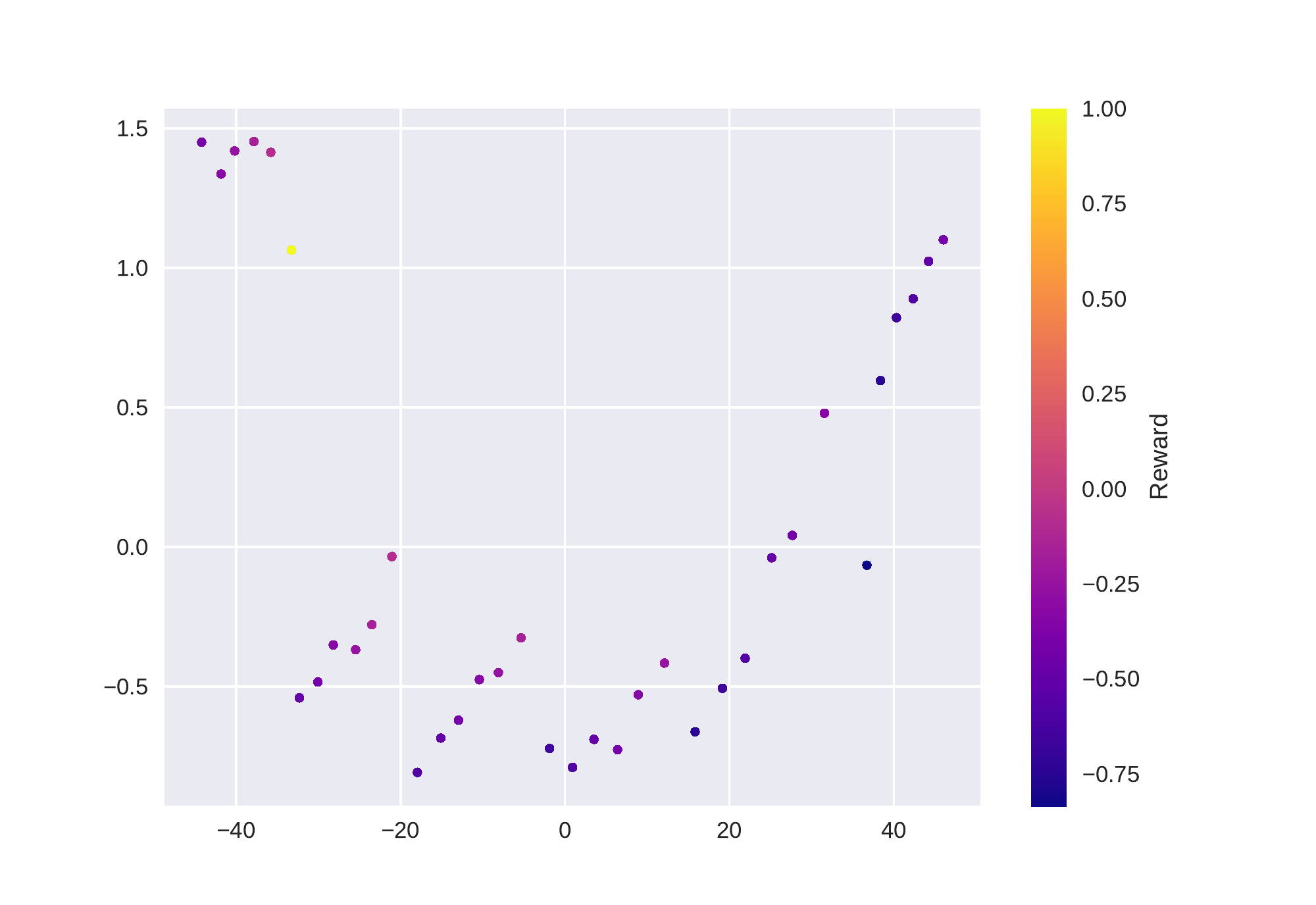}
\captionsetup{justification=centering}
    \caption{JSAE-C \cite{pritz2020joint}}
\label{fig:jsaec}
    \end{subfigure}
\caption{First two principal components of the learned state representations of the $6\times6$ grid-world in Figure \ref{fig:env-1obj}. The color of the state predictions is done accordingly to the reward function in Equation (\ref{reward_function_grid}).}
    \label{fig:state_representation_6x6maze}
    \end{figure*}

The learned representations in the $6 \times 6$ grid-world are shown in Figure \ref{fig:state_representation_6x6maze}.
    
\begin{figure*}[ht!]
\centering
    \begin{subfigure}{0.33\textwidth}
    \centering
\includegraphics[width=1.0\linewidth,page=8]{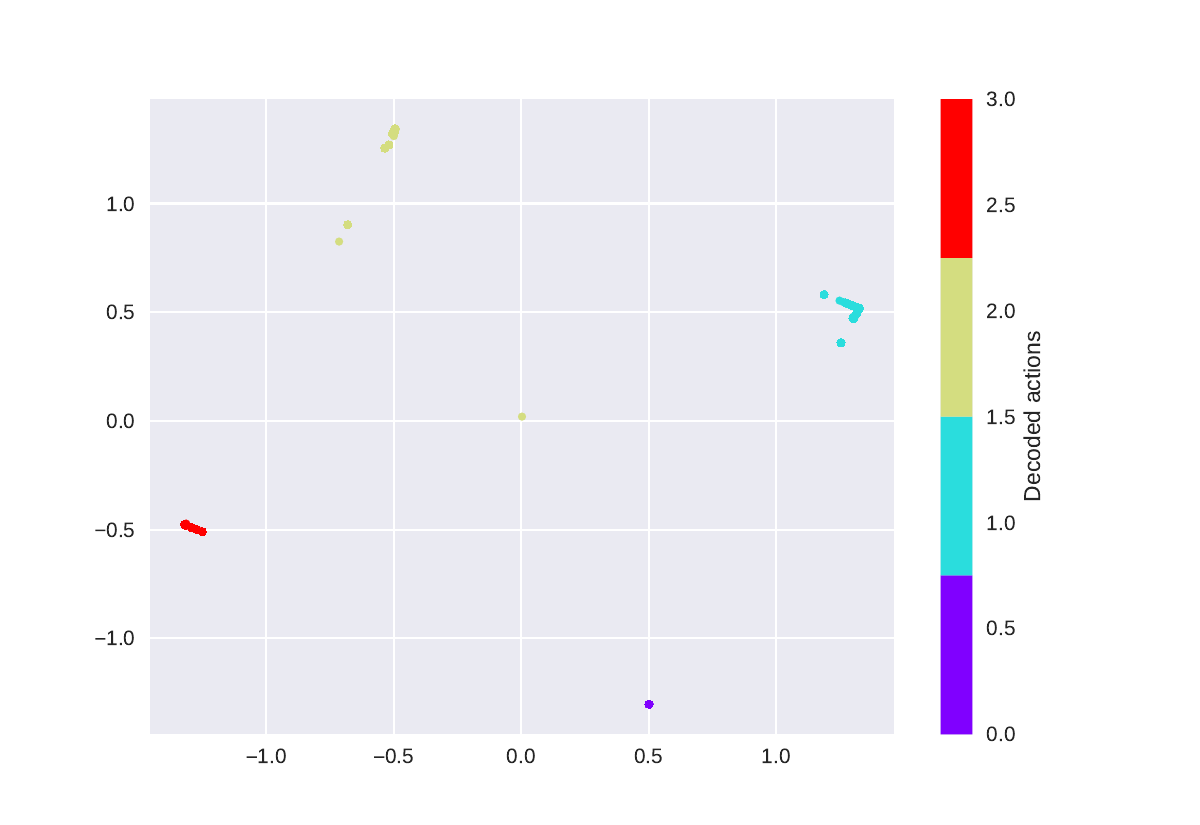}
\captionsetup{justification=centering}
    \caption{Maze $6\times6$, latent actions}
\label{fig:maze6a}
    \end{subfigure} 
        \begin{subfigure}{0.33\textwidth}
        \centering
\includegraphics[width=1.0\linewidth,page=11]{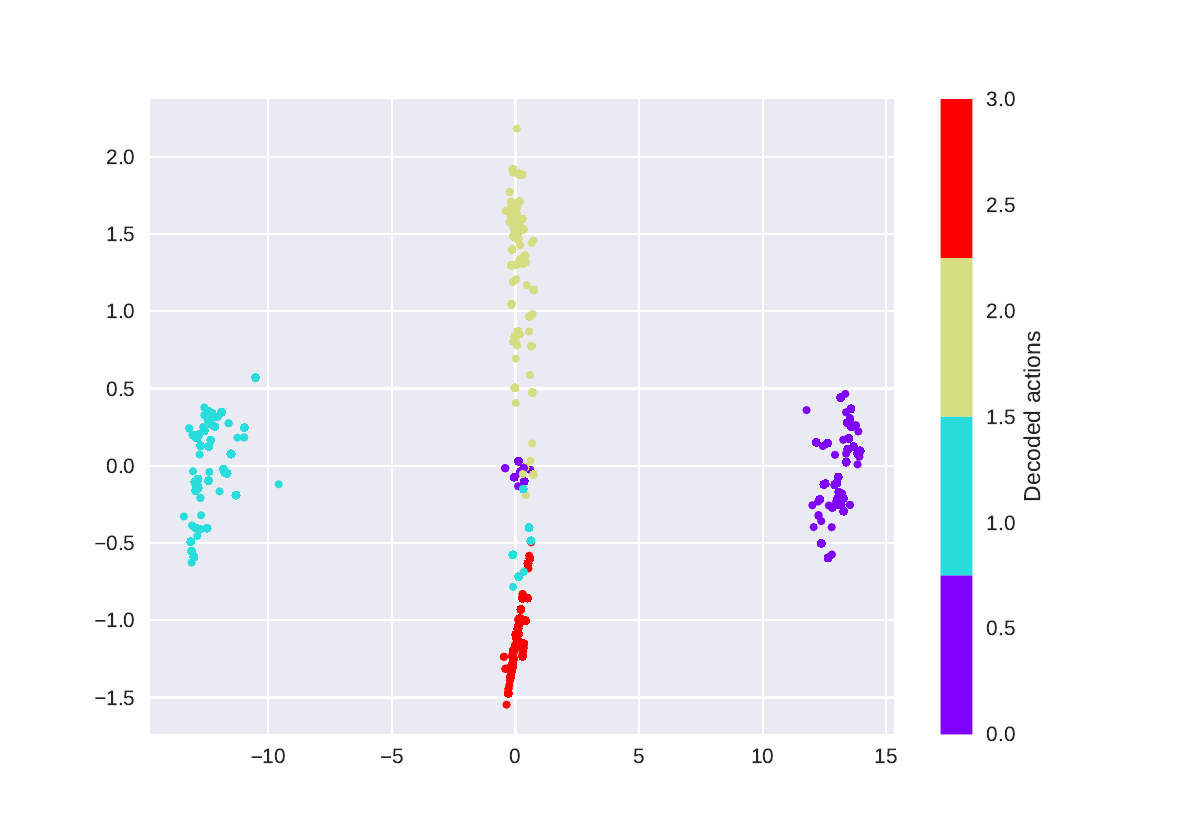}
\captionsetup{justification=centering}
    \caption{Maze $6\times6$, latent transitions}
\label{fig:maze6t}
    \end{subfigure} 
\caption{First two principal components of the learned action space $\mathcal{\bar{A}}$ and the $\Delta \bar{\text{T}}$ in the $6\times6$ grid-worlds in Figure \ref{fig:env-1obj}.}
    \label{fig:grid-envs-latent_act6x6}
    \end{figure*}

The learned action representation and the latent transitions in the $6\times6$ maze are shown in Figure \ref{fig:grid-envs-latent_act6x6}.

\end{document}